\documentclass{article}

\PassOptionsToPackage{numbers,sort&compress}{natbib}

\usepackage[final]{neurips_2024}

\usepackage{shortex}

\newcommand{\polylog}{\mathrm{polylog}}

\newcommand{\oneish}{(\sim\!1)}

\allowdisplaybreaks

\title{General bounds on the quality of Bayesian coresets}

\author{%
  Trevor Campbell\thanks{\url{https://trevorcampbell.me}}\\
  Department of Statistics\\
  University of British Columbia\\
  \texttt{trevor@stat.ubc.ca}\\
}

\newcommand{\KLmin}{\underline{\KL}}
\newcommand{\KLmax}{\overline{\KL}}

\begin{document}

\maketitle

\begin{abstract}
Bayesian coresets speed up posterior inference in the large-scale data regime
by approximating the full-data log-likelihood function with a surrogate
log-likelihood based on a small, weighted subset of the data. But while
Bayesian coresets and methods for construction are applicable in a wide range
of models, existing theoretical analysis of the posterior inferential error
incurred by coreset approximations only apply in restrictive settings---i.e.,
exponential family models, or models with strong log-concavity and smoothness
assumptions. This work presents general upper and lower bounds on the
Kullback-Leibler (KL) divergence of coreset approximations that reflect the
full range of applicability of Bayesian coresets. The lower bounds require only
mild model assumptions typical of Bayesian asymptotic analyses, while the upper
bounds require the log-likelihood functions to satisfy a generalized
subexponentiality criterion that is weaker than conditions used in earlier
work. The lower bounds are applied to obtain fundamental limitations on the
quality of coreset approximations, and to provide a theoretical explanation for
the previously-observed poor empirical performance of importance sampling-based
construction methods. The upper bounds are used to analyze the performance
of recent subsample-optimize methods. The flexibility of the theory is demonstrated in
validation experiments involving multimodal, unidentifiable, heavy-tailed
Bayesian posterior distributions.

\end{abstract}

\section{Introduction}\label{sec:intro}
Large-scale data is now commonplace in scientific and commerical applications
of Bayesian statistics. But despite its prevalence, and the corresponding
wealth of research dedicated to scalable Bayesian inference, there are still
suprisingly few general methods that provably provide inferential results, within some reasonable tolerated error, 
at a significant computational cost savings.  
Exact Markov chain Monte Carlo (MCMC) methods require many full passes over the
data [\citealp[Ch.~6--12]{Robert04}, \citealp[Ch.~11--12]{Gelman13}],
limiting the utility of these methods when even a single pass is expensive.
A wide range of MCMC methods that access only a subset of data per iteration, e.g.,
via delayed acceptance \cite{Christen05,Banterle19,Payne14,Sherlock17},
pseudomarginal or auxiliary variable methods \cite{Doucet15,Maclaurin14,Quiroz21},
and basic subsampling \cite{Welling11,Ahn12,Korattikara14,Chen15},
provide at most a minor improvement over full-data MCMC \cite{Johndrow20,Bardenet17,Nagapetyan17}.
On the other hand, methods including carefully
constructed log-likelihood function control variates can provide substantial gains \cite{Baker19,Nemeth21,Quiroz19}.
However, black-box control variate constructions for large-scale data often rely on assumptions such as
posterior density differentiability and unimodality that do not hold in many popular models, e.g., those with discrete variables or multimodality.
See \cite{Bardenet17,Quiroz18} for a survey of scalable MCMC methods.
Parametric approximations via variational inference \cite{Blei17} or
the Laplace approximation \cite{Shun95,Hall11} can be obtained scalably using stochastic optimization methods,
but existing general theoretical guarantees for these methods again typically rely on posterior normality assumptions
[\citealp[p.~141--144]{vanderVaart00},\citealp{Wang18,CheriefAbdellatif18,Yang18,Alquier20,Xu22,Miller21}]
(see \cite{Blei17,Zhang19} for a review). 

Although many existing methods rely on asymptotic normality or unimodality
in the large-scale data regime, the problem of handling large-scale
data in Bayesian inference does not fundamentally require this structure. 
Instead, one can more generally exploit \emph{redundancy} in the data (i.e., the existence of good approximate sufficient statistics),
which can be used to draw principled conclusions about a large data set based only on a small
fraction of examples. Indeed, while approximate posterior normality often does not hold in models with
latent discrete or combinatorial objects, weakly identifiable or unidentifiable parameters,
persisting heavy tails, multimodality, etc., such models can and regularly do exhibit significant redundancy in the data
that can be exploited for faster large-scale inference. \emph{Bayesian coresets} \cite{Huggins16}---which involve replacing
the full dataset during inference with a sparse weighted subset---are based on this notion of exploiting data redundancy.
Empirical studies have shown the existence of high-quality coreset posterior approximations 
constructed from a small fraction of the data, even in models that violate posterior normality assumptions and for which
standard control variate techniques work poorly \cite{Campbell19JMLR,Campbell19b,Campbell18,Chen22,Naik22}.
However, existing theoretical support for Bayesian coresets in the literature is limited.
There exist no lower bounds on Bayesian coreset approximation error, and
while upper bounds do exist, they currently impose restrictive assumptions. In particular, 
the best available theoretical upper bounds to date 
apply to exponential family models \cite{Chen22,Jankowiak22} and models with strongly log-concave
and locally smooth log-densities \cite{Naik22}.

This article presents new theoretical techniques and results regarding the
quality of Bayesian coreset approximations.  The main results are two general
large-data asymptotic lower bounds on the KL divergence
(\cref{thm:diffable_lower,thm:diffable_lower_min}),
as well as a general upper bound on the KL divergence
(\cref{thm:subexpupper}) under the assumption that the log-likelihoods satisfy
a multivariate generalization  of subexponentiality (\cref{def:subexp}).
The main general results in this paper lead to various 
novel insights about specific Bayesian coreset construction methods. Under mild assumptions,
\bitem
\item common importance-weighted coreset constructions (e.g.~\cite{Huggins16}) require
a coreset size $M$ proportional to the dataset size $N$ (\cref{cor:importanceweighted}),
even with post-hoc optimal weight scaling (\cref{cor:scaledimportanceweighted}), and thus yield
a negligible improvement over full-data inference;
\item \emph{any} construction algorithm requires a coreset size $M > d$
when the log-likelihood function is determined by $d$ parameters locally
around a point of concentration (\cref{cor:highdimlowerbd});
\item subsample-optimize coreset construction algorithms (e.g.~\cite{Jankowiak22,Chen22,Naik22,Chen24}) 
achieve an asymptotically bounded error with a coreset size $\polylog N$ in a wide variety of models (\cref{cor:subopt}).
\eitem
The paper includes empirical validation of the main theoretical claims on 
two models that violate common assumptions made in the literature:
a multimodal, unidentifiable Cauchy location model with a heavy-tailed prior, and an unidentifiable
logistic regression model with a heavy-tailed prior and persisting posterior heavy tails.
Experiments were performed on a computer with an Intel Core i7-8700K and 32GB of RAM.
Proofs of all theoretical results
 may be found in \cref{sec:proofs}.

\textbf{Notation.} We use standard asymptotic growth symbols $O, \Omega, \Theta, o, \omega$ (see, e.g., \cite[Sec.~3.3]{Cormen22}),
and their probabilistic variants $O_p, \Omega_p, \Theta_p, o_p, \omega_p$ (see, e.g., \cite[Sec.~2.2]{vanderVaart00}).
We use the same symbol to denote a measure $\pi$ and its density $\pi(\cdot)$ with respect to a specified dominating measure.
We also regularly suppress integration variables and differential symbols in integrals throughout for notational brevity when these are clear from context;
for example, $\int \pi \exp(\ell)$ is shorthand for $\int \pi(\d\theta) \exp(\ell(\theta))$.
Finally, the pushforward of a measure $\pi$ by a map $\eta$ is denoted simply $\eta\pi$.

\section{Background}\label{sec:background}
Define a target probability distribution $\pi$ on a space $\Theta$
comprised of a sum of $N$ potentials $\ell_n : \Theta \to \reals$, $n=1,\dots, N$
and a base distribution $\pi_0(\d\theta)$,
\[
\pi(\d\theta) &= \frac{1}{Z} \exp\left(\ell(\theta)\right) \pi_0(\d\theta), 
&
\ell(\theta) &= \sum_{n=1}^N \ell_n(\theta), 
&
\theta&\in\Theta,\label{eq:coreset_base_model}
\]
where the normalization constant $Z$ is not known. 
In the Bayesian context,
this distribution corresponds to a Bayesian posterior distribution
for a statistical model with prior $\pi_0$ and
conditionally i.i.d.~data $X_n$, where $\ell_n(\theta) = \log p(X_n | \theta)$. 
The goal is to compute or approximate expectations under $\pi$; but the likelihood
$\ell$ (and its gradient) becomes expensive to evaluate when $N$ is large. 
To avoid this cost,
\emph{Bayesian coresets} \cite{Huggins16,Campbell19JMLR,Campbell19b,Campbell18,Chen22,Naik22}
involve replacing the target with a surrogate density 
\[
\pi_w(\d\theta) &= \frac{1}{Z(w)}\exp\left(\ell_w(\theta)\right) \pi_0(\d\theta), 
&
\ell_w(\theta) &= \sum_{n=1}^N w_n \ell_n(\theta),
&
\theta\in\Theta,\label{eq:coreset}
\]
where $w\in\reals^N$, $w\geq 0$ are a set of weights,
and $Z(w)$ is the new normalizing constant. 
If $w$ has at most $M \ll N$ nonzeros, the $O(M)$ cost of evaluating
$\sum_n w_n \ell_n$ (and its gradient) is a significant improvement upon
the original $O(N)$ cost.
In this work, the problem of coreset construction is formulated in the
data-asymptotic limit; a coreset construction method should
\bitem
\item run in $o(N)$ time and memory (or at most $O(N)$ with a small leading constant),
\item produce a small coreset of size $M = o(N)$,
\item produce a coreset with $O(1)$ posterior forward/reverse KL divergence as $N\to\infty$.
\eitem
These three desiderata ensure that the effort spent constructing and sampling from the coreset posterior 
is worthwhile: the coreset provides a meaningful reduction in computational cost compared with 
standard Markov chain Monte Carlo algorithms, and has a bounded approximation error.

\section{Lower bounds on approximation error} \label{sec:lower_bounds}
This section presents lower bounds on the KL divergence of coreset approximations for general models and data generating processes.
The first key steps in the analysis are to write all expectations in terms of distributions that do not depend on $w$,
and to remove the difficult-to-control influence of the tails of $\pi$ and $\pi_w$ by restricting certain integrals to some small subset $B\subseteq \Theta$ 
of the parameter space. \cref{lem:kl_lower}, the key theoretical tool used in this section, achieves both of these two goals; note that the result has
no major assumptions and applies generally in any setting that a Bayesian coreset can be used.
For convenience, define
\[
\KLmin(w) := \min\{\KL(\pi_w||\pi),\KL(\pi||\pi_w)\},
\]%
and the decreasing, nonnegative function $f:\reals_+ \to \reals_+$,
\[
f(x) = \lt\{ 
\begin{array}{ll} -\log x + x - 1 & 0\leq x \leq 1 \\ 
0 & x > 1.
\end{array}\rt. 
\]
\vspace{-.2cm}
\blem[Basic KL Lower Bound]\label{lem:kl_lower}
For all measurable $B\subseteq \Theta$ and coreset weights $w$,
\[
\KLmin(w) \geq f(J_B(w)) \geq 0,
\]
where
\[
J_B(w) &= \frac{\int_B\pi_0\exp\frac{1}{2}(\ell + \ell_w)}{\sqrt{\int \pi_0\exp(\ell) \int\pi_0\exp(\ell_w)}} + \sqrt{\pi(B^c)}.
\]
\elem
Note that while the integrals in the fraction denominator in $J_B(w)$ range over the whole $\Theta$ space, a further lower bound
on $\KLmin(w)$ can be obtained by restricting their domains arbitrarily.
Also, crucially, the bound in \cref{lem:kl_lower} does not depend on $\pi_w(B^c)$, which would be difficult to analyze
without detailed knowledge of the tail behaviour of $\pi_w$ as a function of the coreset weights $w$. 
Although the bound in \cref{lem:kl_lower} applies generally, it is most useful when 
$B$ is small (so that simple local approximations of $\ell$ and $\ell_w$ can be used),
$\pi$ concentrates on $B$ (so that $\pi(B^c) \approx 0$),
and $\pi$ and $\pi_w$ are very different when restricted to $B$; 
the behaviour of the bound in this case is roughly (see the proof in \cref{sec:proofs})
$f(J_B(w)) \approx -\log(1-\TV(\pi,\pi_w))$.
Finally, note that \cref{lem:kl_lower} remains valid if one replaces $\ell_w$ with $\ell_w-c$ and 
$\ell$ with $\ell-c'$ for any constants $c,c'$ that do not depend on $\theta$
but may depend on the data and coreset weights $w$.

For the remainder of this section, consider the setting where $\Theta$ is a measurable subset of $\reals^d$ 
for some $d\in\nats$, fix some $\theta_0 \in \Theta$, and assume each $\ell_n$ is differentiable in a neighbourhood of $\theta_0$. Let
\[
\sbw = \sum_n w_n \qquad g = \grad\ell(\theta_0) \qquad g_w = \grad\ell_w(\theta_0).
\]
\cref{thm:diffable_lower,thm:diffable_lower_min} characterize KL divergence lower bounds in terms of 
the sum of the coreset weights $\sbw$ and the log-likelihood gradients $g, g_w$.
Intuitively for the full data set where all $w_n=1$ and $\sbw=N$, 
and an \iid data generating process from the likelihood with parameter $\theta_0$,
the central limit theorem asserts under mild conditions that $g_w/\sbw \convp 0$
at a rate of $N^{-1/2}$. \cref{thm:diffable_lower,thm:diffable_lower_min} below provide KL lower bounds when the coreset construction
algorithm does not match this behavior. In particular, \cref{thm:diffable_lower} provides results that are useful when 
$g_w/\sbw \convp 0$ occurs reasonably quickly but slower than $N^{-1/2}$, while \cref{thm:diffable_lower_min} strengthens the conclusion when 
$g_w/\sbw \convp 0$ very slowly or not at all.
The major benefit of \cref{thm:diffable_lower,thm:diffable_lower_min}  for analyzing coreset construction methods 
is that they reduce the problem of analyzing posterior KL divergence 
to the much easier problem of analyzing the 2-norm $\|\cdot\|_2$ of a weighted sum of random vectors in $\reals^d$.

Consider a sequence $r\to 0$ as $N\to\infty$, and for a fixed matrix $H\succ 0$ let 
\[
	B = \{\theta : (\theta-\theta_0)^TH(\theta-\theta_0)\leq r^2\}
\]
be a sequence of neighbourhoods around $\theta_0$;
these will appear in \cref{assum:lower,assum:lower2,thm:diffable_lower,thm:diffable_lower_min} below. 
Note that throughout, all asymptotics will be taken as $N\to\infty$,
and various sequences (e.g., $r$ and $B$) are implicitly indexed by $N$.
To simplify notation, this dependence is suppressed.
\cref{assum:lower} makes some weak assumptions about the model and data generating process: it 
intuitively asserts that the potential functions are sufficiently smooth around $\theta_0$,
that $r\to 0$ slowly, and that $\pi$ concentrates at $\theta_0$ at a usual rate. Note 
that \cref{assum:lower} does not assume data are generated \iid and 
places no conditions on the coreset construction algorithm.
\bassum\label{assum:lower}
$\pi_0$ has a density with respect to the Lebesgue measure, $\pi_0(\theta_0) > 0$,
each $\ell_n(\theta)$ and $\pi_0(\theta)$ are twice differentiable in $B$ for sufficiently large $N$, 
and
\[
\sup_{\theta\in B} \lt\|-\frac{1}{N}\grad^2\ell(\theta) - H\rt\|_2 = o_p(1), 
\quad \lt\|\frac{g}{N}\rt\|_2 = O_p\lt(N^{-1/2}\rt), \quad \quad Nr^2 = \omega(1).
\]
\eassum
Two additional assumptions related to the coreset construction algorithm---namely, that it works
well enough that $\frac{1}{\sbw}\sum_n w_n \grad^2\ell_n(\theta) \convp H$ and $g_w/\sbw \convp 0$
at a rate faster than $r\to 0$---lead to
asymptotic lower bounds on the best possible quality of coresets produced by the algorithm, as well as lower
bounds even after optimal post-hoc scaling of the weights.
\bthm\label{thm:diffable_lower}
Suppose \cref{assum:lower} holds. If
\[
\sup_{\theta\in B}\lt\|-\frac{1}{\sbw}\grad^2\ell_w(\theta) - H\rt\|_2 = o_p(1), \quad \lt\|\frac{g_w}{\sbw}\rt\|_2 = o_p(r),
\]
then
\[
\KLmin(w) &\!\geq \!
O_p(1) \!+\!\Omega_p(1)\min\lt\{\!-\log \pi(B^c),
\frac{N\sbw}{N+\sbw}\lt\|\frac{g}{N}-\frac{g_w}{\sbw}\rt\|_2^2
\!+\! d\log\frac{(N+\sbw)^2}{N\max\{\sbw, 1/r^2\}}\rt\}\\
\min_{\alpha \geq 0}\KLmin(\alpha w)&\!\geq\! O_p(1) \!+\! \Omega_p(1)\min\lt\{\!-\log \pi(B^c), d\log\left(N\left\|\frac{g}{N}-\frac{g_w}{\sbw}\right\|_2^2\right)\rt\}.
\]
\ethm
\cref{thm:diffable_lower} is restricted to the case where the coreset algorithm is performing reasonably well.
\cref{thm:diffable_lower_min} extends the bounds to the case where the algorithm is performing poorly, in the sense that it is 
unable to make $\frac{g_w}{\sbw}\convp 0$ at a rate faster than $r\to 0$
(or perhaps $\frac{g_w}{\sbw}$ does not converge to 0 at all). In order to draw conclusions in this setting,
we need a weak global assumption on the potential functions. A function $f : \Theta \to \reals$ is \emph{$L$-smooth below at $\theta_0$} if 
\[
\forall \theta \in \Theta, \quad f(\theta) \geq f(\theta_0) + \grad f(\theta_0)^T(\theta-\theta_0) - \frac{L}{2}\lt\|\theta-\theta_0\rt\|_2^2.\label{eq:Lsmoothbelow}
\]
Note that $L$-smoothness below is weaker than Lipschitz smoothness and does not imply concavity; \cref{eq:Lsmoothbelow} 
restricts the growth of the function only in the negative direction, and only when the expansion is taken at $\theta_0$. 
\cref{assum:lower2} asserts that the potential functions are smooth below.
\bassum\label{assum:lower2}
There exist $L_0, \dots, L_N, L > 0$
such that $\log\pi_0$ is $L^2_0$-smooth below at $\theta_0$,
for each $n\in[N]$ $\ell_n$ is $L^2_n$-smooth below at $\theta_0$,
and $\frac{1}{N}\sum_{n=1}^N L_n^2 \convp L^2$.
\eassum
\cref{thm:diffable_lower_min} uses
\cref{assum:lower,assum:lower2} and additional assumptions related to the coreset construction algorithm to
 obtain lower bounds in a setting that relaxes the ``performance'' conditions in \cref{thm:diffable_lower}:
$-\frac{1}{\sbw}\sum_n w_n\grad^2\ell_n(\theta)$ no longer needs to converge to $H$ in probability, and $g_w/\sbw$ can converge to 0 slowly or not at all.
\bthm\label{thm:diffable_lower_min}
Suppose \cref{assum:lower,assum:lower2} hold. If there exist $\alpha,\beta > 0$ such that
\[
\P\lt(\forall\theta \in B, \,\,-\frac{1}{\sbw}\grad^2\ell_w(\theta) \succeq \alpha H\rt) \to 1, \quad \P\lt(\frac{1}{\sbw}\sum_n w_n L^2_n \leq \beta L^2\rt)\to 1,\quad \lt\|\frac{g_w}{\sbw}\rt\| = \omega_p(r),
\]
then
\[
\KLmin(w) \geq  O_p(1) + \Omega_p(1)\min\lt\{-\log\pi(B^c), d\log\lt(N\min\lt\{\lt\|\frac{g_w}{\sbw}\rt\|^2, 1\rt\}\rt)\rt\}.
\]
\ethm

An important final note in this section is that while
\cref{thm:diffable_lower,thm:diffable_lower_min}, as stated, require choosing
$\Theta$ to be some measurable subset of $\reals^d$ and that the posterior $\pi$ concentrates around some point
of interest $\theta_0\in\reals^d$, these results can be generalized to a wider
class of models and spaces.  In particular, \cref{cor:lower_other} demonstrates
that if $\Theta$ is arbitrary, but the potential functions $\ell_n$ only depend on $\theta$ through some
other function $\eta : \Theta \to \reals^d$, that the conclusions of
\cref{thm:diffable_lower,thm:diffable_lower_min} still hold.

\bcor\label{cor:lower_other}
Suppose $\Theta$ is an arbitrary measurable space,
and the potential functions take the form $\ell_n(\eta(\theta))$
for some measurable function $\eta : \Theta \to \reals^d$.
Then if the assumptions of 
\cref{thm:diffable_lower,thm:diffable_lower_min} hold for 
potentials $(\ell_n)_{n=1}^N$ as functions on $\reals^d$ and
pushforward prior $\eta\pi_0$ on $\reals^d$,
the stated lower bounds also hold for $\min\{\KL(\pi||\pi_w), \KL(\pi_w||\pi)\}$.
\ecor

\section{Lower bound applications}\label{sec:applications_lower}

In this section, the general theoretical results from \cref{sec:lower_bounds}
are applied to specific algorithms, Bayesian models, and data generating
processes to explain previously observed empirical behaviour of coreset
construction, as well as to place fundamental limits on the necessary size of
coresets. Consider a setting where the data
$X_n$ arise as an \iid sequence drawn from some probability distribution $\nu$,
$\ell_n(\eta(\theta)) = \log p(X_n | \eta(\theta))$ for $\eta : \Theta \to \reals^d$, $\eta_0 = \eta(\theta_0)$, 
and the following technical criteria hold (where $\E$ denotes expectation under the data generating process):
\bitem
\item[(A1)] $\E\left[\grad\ell_n(\eta_0)\right] = 0$ and $H=\E\left[-\grad^2\ell_n(\eta_0)\right] = \E\left[\grad\ell_n(\eta_0)\grad\ell_n(\eta_0)^T\right] \succ 0$.
\item[(A2)] $\E\left[\|\grad\ell_n(\eta_0)\|_2^{2+\delta}\right] < \infty$ for some $\delta>0$ and $\E\left[\|\grad^2\ell_n(\eta_0)\|^2_F\right] < \infty$.
\item[(A3)] On a neighbourhood of $\eta_0$, $\|\grad^2\ell_n(\eta) - \grad^2\ell_n(\eta_0)\|_2 \leq R(X_n)\|\eta-\eta_0\|_2$, $\E\left[R(X_n)\right] < \infty$.
\item[(A4)] $\eta\pi_0$ is twice differentiable a neighbourhood of $\eta_0$, and $\pi(\eta_0) > 0$.
\item[(A5)] For all $r\to 0$ such that $r^2=\omega\lt(\log N/N\rt)$, $-\log\eta\pi\lt(\|\eta-\eta_0\| > r\rt) = \Omega_p(Nr^2)$.
\eitem

These conditions apply to a wide range of models, e.g., 
an unidentifiable, multimodal location model posterior with heavy tails on $\Theta = \reals$,
where the Bayesian model is specified by
\[
\theta &\dist \Cauchy(0, 1) & 
(X_n)_{n=1}^N &\distiid \Cauchy(\theta^2, 1), \label{eq:cauchymodel}
\]
and the data are generated from the likelihood with parameter $\theta_0 = 5$,
and an unidentifiable logistic regression posterior with heavy tails on $\reals^2$,
where the Bayesian model is specified by
\[
\theta &\dist \Cauchy(0, I) & 
Y_n &\distind \Bern\lt(\frac{1}{1+e^{-X_n^TA\theta}}\rt) &
A &= \bbmat 1 & 1\\ 1 & 1\ebmat, \label{eq:logregmodel}
\]
the covariates are generated via $X_n \distiid \Unif(\{x\in\reals^2 : \|x\|_2\leq 1\})$,
and the observations $Y_n$ are generated from the likelihood with parameter $\theta_0 = \bbmat 1 & 6 \ebmat^T$.
See \cref{prop:cauchylogrega15} in \cref{sec:proofs} for the verification of (A1-5) for these two models.
Example posterior log-densities for these models are displayed in \cref{fig:models}.

\bfig
\bsubfig{0.38\textwidth}
\centering\includegraphics[width=\textwidth]{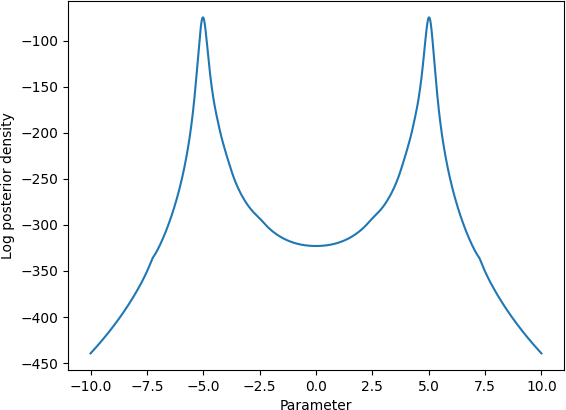}
\caption{}\label{fig:cauchymodel}
\esubfig
\hfill\bsubfig{0.29\textwidth}
\centering\includegraphics[width=\textwidth]{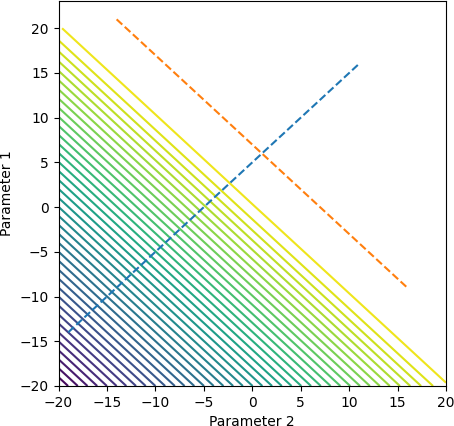}
\caption{}\label{fig:logregmodel}
\esubfig
\hfill\bsubfig{0.30\textwidth}
\centering\includegraphics[width=0.8\textwidth]{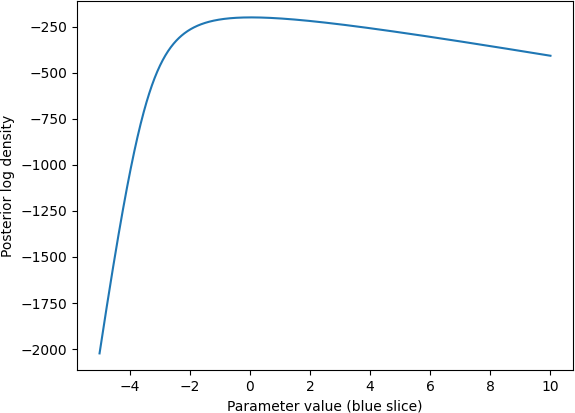}
\centering\includegraphics[width=0.8\textwidth]{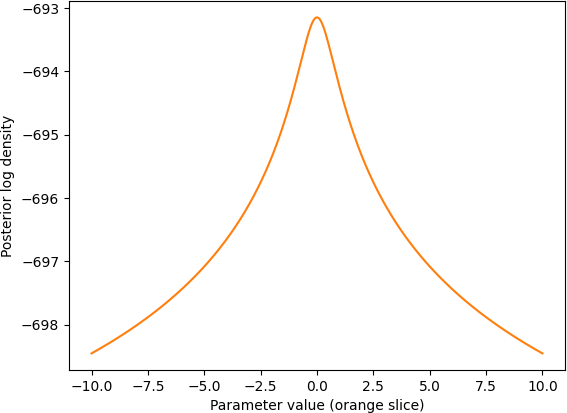}
\esubfig
\caption{Example unnormalized posterior densities given 50 data points for (\ref{fig:cauchymodel}) the Cauchy location model and
(\ref{fig:logregmodel}) the logistic regression model. The orange and blue dashed lines in (\ref{fig:logregmodel})
indicate one-dimensional slices that are shown in the rightmost panels.}\label{fig:models}
\efig

\begin{algorithm}[t!]
\caption{Importance-weighted coreset construction}\label{alg:subsampling}
\begin{algorithmic}
\State Compute probabilities $(p_n)_{n=1}^N$ (may depend on the data and model)
\State Draw $I_1, \dots, I_M \distiid \Cat(p_1, \dots, p_N)$
\State For each $n$, set $w_n = \frac{1}{Mp_n}\sum_{m=1}^M \1[I_m = n]$.
\State \Return $(w_n)_{n=1}^N$
\end{algorithmic}
\end{algorithm}
\begin{algorithm}[t!]
\caption{Scaled importance-weighted coreset construction}\label{alg:scaledsubsampling}
\begin{algorithmic}
\State Obtain coreset weights $(w_n)_{n=1}^N$ via \cref{alg:subsampling}
\State Compute $\alpha^\star = \argmin_{\alpha \geq 0} \KL(\pi_{\alpha w}||\pi)$
\State \Return $(\alpha^\star w_n)_{n=1}^N$
\end{algorithmic}
\end{algorithm}

\subsection{Minimum coreset size for importance-weighted coresets}

A popular algorithm for coreset construction that has appeared in a wide
variety of domains---e.g., Bayesian 
inference [\citealp{Huggins16}, \citealp[Section 4.1]{Campbell19JMLR}], frequentist 
inference (e.g., \cite{Ma15,Wang18b,Wang19,Ai21,Wang21}), and optimization (see \cite{Feldman20} for a recent survey)---involves subsampling of the
data followed by an importance-weighting correction.  
The pseudocode is given in \cref{alg:subsampling}.
Note that $\E[w_n] = 1$, and so $\E[\ell_w] = \ell$; the coreset
potential is an unbiased estimate of the exact potential. 
The advantage of this method is that it is straightforward and computationally efficient.
If the sampling probabilities are uniform $p_n = \nicefrac{1}{N}$, then
\cref{alg:subsampling} constructs a coreset in $O(M)$ time and $O(M)$ memory.
Nonuniform probabilities $p_n$ require $\Omega(N)$ time,
as they require a pass over all $N$ data points to compute each $p_n$ \cite{Huggins16,Wang18b}
followed by sampling the coreset, e.g., via an alias table \cite{Walker74,Walker77}. 
However, empirical results produced by this methodology have generally been underwhelming, even with
carefully chosen sampling probabilities; see, e.g., Figure 2 of \cite{Huggins16}.

\cref{cor:importanceweighted} explains these poor results: Bayesian coresets 
constructed via \cref{alg:subsampling} must satisfy $M\propto N$ in order to maintain a bounded $\KLmin(w)$
in the data-asymptotic limit. In other words, such coresets do not satisfy the desiderata in \cref{sec:background}.
The only restriction is that there exist constants $c, C > 0$ such 
that for all $N\in\nats$, the sampling probabilities $(p_n)_{n=1}^N$ satisfy
\[
\text{(A6) } \qquad0 < c \leq \min_n N p_n \leq \max_n Np_n \leq C < \infty \quad a.s. \label{eq:pncondition}
\]
The lower threshold ensures that the variance of the importance-weighted log-likelihood is not too large, while the upper threshold
ensures sufficient diversity in the draws from subsampling. The condition in \cref{eq:pncondition} is not a major 
restriction, in the sense that performance should deteriorate even further when it does not hold.
The  $(p_n)_{n=1}^N$ may otherwise depend arbitrarily on the data and model.
\bcor\label{cor:importanceweighted}
Given (A1-6), $M\to\infty$, and $M = o(N)$, coresets produced by \cref{alg:subsampling} satisfy
\[
\KLmin(w) = \Omega_p\left(\frac{N}{M}\right). \label{eq:NoverMresult}
\]
\ecor
The intuition behind \cref{cor:importanceweighted} is that both the true posterior
and the importance-weighted coreset posterior are asymptotically approximately normal with variance
$\propto 1/N$ as $N\to\infty$; 
however, the coreset posterior mean is roughly $\propto M^{-1/2}$
away from the posterior mean, because the subsample is of size $M$. 
The KL divergence between two Gaussians
is lower-bounded by the inverse variance times the mean difference squared, yielding $\approx N/M$
as in \cref{eq:NoverMresult}.

Given the intuition that the coreset posterior mean is far from the posterior mean relative to their variances,
it is worth asking whether one can apply a small amount of effort to ``correct'' the importance-weighted coreset 
by scaling the weights (and hence the variance) down,
as shown in \cref{alg:scaledsubsampling}. Unfortunately, \cref{cor:scaledimportanceweighted} demonstrates that even with optimal scaling,
$M\propto N$ is still required in order to maintain a bounded KL divergence as $N\to\infty$.
\bcor\label{cor:scaledimportanceweighted}
Given (A1-6), $M\to\infty$, and $M = o(N)$, coresets produced by \cref{alg:subsampling} satisfy
\[
\min_{\alpha > 0} \KLmin(\alpha w) = \Omega_p\left(\log\frac{N}{M}\right).
\]
\ecor

\cref{fig:importanceweighted} provides empirical confirmation of
\cref{cor:importanceweighted,cor:scaledimportanceweighted} on the Cauchy
location and logistic regression models  in
\cref{eq:cauchymodel,eq:logregmodel}. In particular, these figures show that
the empirical rates of growth of KL as a function of $N$ closely matches
$\Omega_p(\frac{N}{M})$ for importance-weighted coresets, and
$\Omega_p(\log\frac{N}{M})$ for the same with post-hoc scaling, for a wide
range of coreset sizes $M \in \{\log N, \sqrt{N}, \nicefrac{1}{2}N\}$. Thus,
importance weighted coreset construction methods do not satisfy the desiderata
in \cref{sec:background} for a wide range of models, and alternate methods should be considered.

\bfig
\centering\includegraphics[width=.48\textwidth]{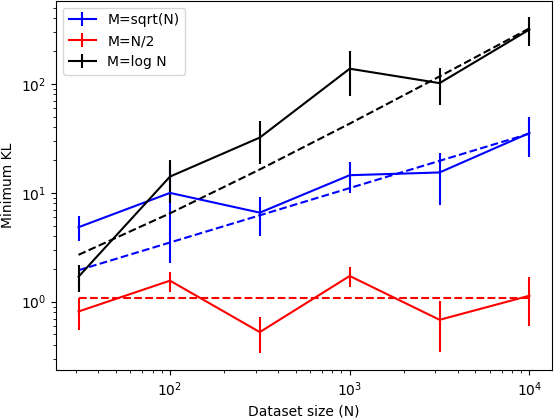}
\hfill
\centering\includegraphics[width=.48\textwidth]{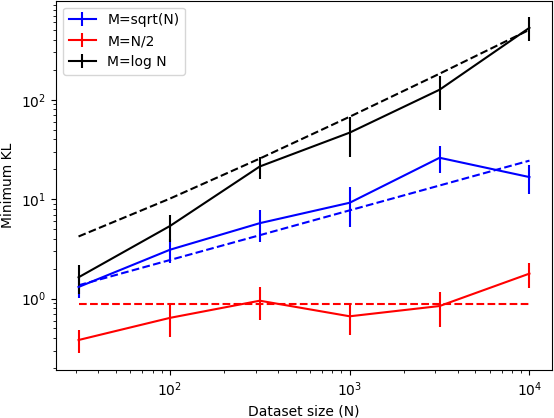}
\\
\centering\includegraphics[width=.48\textwidth]{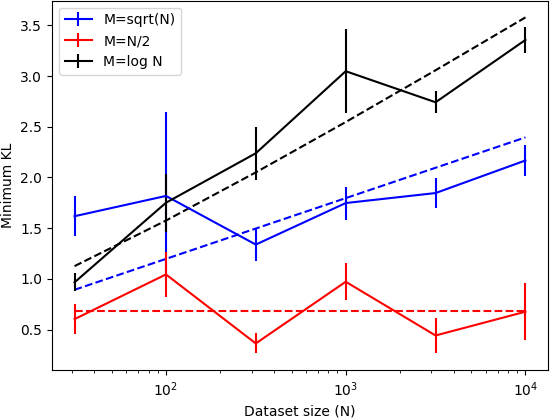}
\hfill
\centering\includegraphics[width=.47\textwidth]{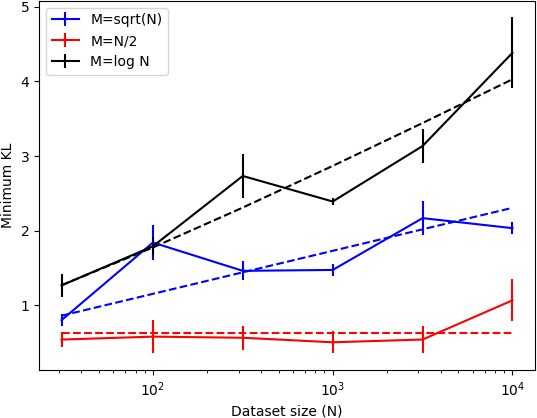}
\caption{Importance-weighted coreset quality, showing the minimum of the forward and reverse KL divergences on the vertical axis
as a function of dataset size $N$ for 3 coreset sizes: $\log N$ (black), $\sqrt{N}$ (blue), and $\nicefrac{1}{2}N$ (red).
Dashed lines indicate predictions from the theory in \cref{cor:importanceweighted,cor:scaledimportanceweighted},
solid lines indicate the mean over 10 trials, and error bars indicate standard error.
The top row shows the quality of basic importance-weighted coresets (note that both horizontal and vertical axes are in log scale), 
while the bottom row shows the quality with optimal post-hoc scaling (note that only the horizontal axis is in log scale).
The left column corresponds to the Cauchy location model, while 
the right column corresponds to the logistic regression model.
Sampling probabilities $p_n$ for both models are set proportional to $X_n^2$, 
thresholded to lie between $0.1/N$ and $10/N$.
}\label{fig:importanceweighted}
\efig

\subsection{Minimum coreset size for any coreset construction}
This section extends the minimum coreset size results from importance-weighted schemes to
\emph{any} coreset construction algorithm. In particular, \cref{cor:highdimlowerbd} shows 
that under (A7)---a strengthening of (A3) and \cref{assum:lower2}---and (A8)---which 
asserts that $\grad\ell_1(\eta_0),\dots,\grad\ell_M(\eta_0)$ are linearly independent \as
and satisfy a technical moment condition---at least $d$ coreset points are required to keep
the KL divergence bounded as $N\to\infty$.
\bitem
\item[(A7)] \cref{assum:lower2} holds and there exists $\gamma > 0$ such that for all sufficiently large $N\in\nats$, 
\[
\forall \eta\in B, n\in[N], \quad  -\grad^2\ell_n(\eta) \succeq \gamma H \quad\text{and}\quad  L^2_n < \gamma^{-1} L^2.
\]
\item[(A8)] For all coreset sizes $M < d$, there exists a $\delta >0$ such that
\[
\E\lt[\lt(1^T(G^TG)^{-1}1\rt)^{M+\delta}\rt] < \infty \qquad G = \bbmat \grad\ell_1(\eta_0) & \dots & \grad\ell_M(\eta_0)\ebmat\in\reals^{d\times M}.
\]
\eitem
\bcor\label{cor:highdimlowerbd}
For a fixed coreset size $M< d$, given (A1-5,7,8),
\[
\min_{w \in\reals_+^N : \|w\|_0 \leq M} \KLmin(w) &= \Omega_p\lt(\log N\rt).
\]
\ecor

\section{Upper bounds on approximation error}\label{sec:upper_bounds}

This section presents upper bounds on the KL divergence of coreset approximations.
As in \cref{sec:lower_bounds}, the first step is to write all expectations in terms of distributions that
do not depend on $w$. \cref{lem:kl_upper} does so without imposing any major assumptions; the result again 
applies generally in any setting that a Bayesian coreset can be used.
For convenience, define
\[
\KLmax(w) := \max\{\KL(\pi_w||\pi), \KL(\pi||\pi_w)\}.
\]
\vspace{-.5cm}
\blem[Basic KL Upper Bound]\label{lem:kl_upper} 
For all coreset weights $w$,
\[
\KLmax(w) &\leq \inf_{\lambda > 0} \frac{1}{\lambda}\log \int \pi \exp\lt((1+\lambda)(\bar\ell_w-\bar\ell)\rt),
\]
where for all $n\in[N]$, $\bar\ell_n = \ell_n - \int \pi \ell_n$, $\bar\ell = \sum_n \bar\ell_n$, and $\bar\ell_w = \sum_n w_n \bar\ell_n$.
\elem
The upper bound in \cref{lem:kl_upper} is nonvacuous (i.e., finite) as long as there exists a $\alpha > 1$ such 
that the $\alpha$ \Renyi divergence $D_\alpha(\pi_w||\pi)$ \cite[p.~3799]{vanErven14} is finite.
Note that as in \cref{lem:kl_lower}, the bound in \cref{lem:kl_upper} remains valid if one replaces $\ell_w$ with $\ell_w-c$ and 
$\ell$ with $\ell-c'$ for any constants $c,c'$ that do not depend on $\theta$ but may depend on the coreset weights $w$ and data.

More practical bounds necessitate an assumption about the behaviour of the potentials $(\ell_n)_{n=1}^N$.
\cref{def:subexp} below asserts that the multivariate moment generating function of $(\ell_n)_{n=1}^N$ is bounded when
the vector is close to 0. This definition is a generalization of the usual definition of subexponentiality for
the univariate setting (e.g., \cite[Sec.~2.7]{Vershynin20}).
\cref{thm:subexpupper} subsequently shows that \cref{def:subexp} is sufficient to obtain simple bounds on $\KLmax$.

\bdefn\label{def:subexp}
For $A\in\reals^{N\times N}$, $A\succeq 0$, and monotone function $h : \reals_+ \to \reals_+$ such that $\lim_{x\to 0} h(x)=h(0) = 0$, 
the potentials $(\ell_n)_{n=1}^N$ are \emph{$(h,A)$-subexponential} if
\[
\forall w\in\reals^N : w^T Aw \leq 1,\qquad
\int \pi \exp\lt(\bar\ell_w\rt) &\leq \exp\lt(h(w^T A w)\rt).
\]
\edefn
\bthm\label{thm:subexpupper}
If the potentials $(\ell_n)_{n=1}^N$ are $(h,A)$-subexponential, then
\[
\forall w\in\reals_+^N : 4(w-1)^TA(w-1)\leq 1, \qquad \KLmax(w) \leq h(4(w-1)^TA(w-1)).
\]
\ethm
\cref{def:subexp}, the key assumption in \cref{thm:subexpupper}, is satisfied by a wide range of models
when choosing $h(x) = x$ and $A \propto \Cov_\pi\lt((\ell_n)_{n=1}^N\rt)$, as demonstrated by \cref{prop:allsubexp}.
Because this case applies widely, let \emph{$A$-subexponential} be shorthand for $(h,A)$-subexponentiality with $h(x)=x$.
\bprop\label{prop:allsubexp}
If for all $w$ in a ball centered at the origin, $\int\pi\exp(\bar\ell_w) < \infty$,
then there exists $\beta > 0$ such that the potentials $(\ell_n)_{n=1}^N$ 
are $\beta \Cov_\pi\lt((\ell_n)_{n=1}^N\rt)$-subexponential.
\eprop
In other words, intuitively, if a coreset construction algorithm produces weights such that $\Var_\pi(\bar\ell_w-\bar\ell)$
is small, then $\KLmax(w)$ is also small. That being said, the generality of \cref{def:subexp} to allow arbitrary $h,A$
is still helpful in obtaining upper bounds in specific cases; see, e.g., \cref{prop:upperrecovered,prop:upperrecovered2}.

\section{Upper bound application: subsample-optimize coresets}\label{sec:applications_upper}

A strategy to construct Bayesian coresets that has recently emerged in the literature, shown in \cref{alg:opt},
is to first subsample the data to select $M$ data points, and  
subsequently optimize the weights for those selected data points \cite{Chen22,Jankowiak22,Naik22}.
The subsampling step serves to pick a reasonably flexible basis of 
log-likelihood functions for coreset approximation, and avoids the slow greedy selection routines 
from earlier work \cite{Campbell19JMLR,Campbell18,Campbell19b}.
The optimization step tunes the weights for the selected basis, 
avoiding the poor approximations of importance-weighting methods.
Indeed, \cref{alg:opt} creates exact coresets $\pi_{w^\star}=\pi$ with high probability in 
Gaussian location models \citep[Prop.~3.1]{Chen22} and finite-dimensional exponential family models \citep[Thm.~4.1]{Naik22}, 
and near-exact coresets with high probability in strongly log-concave models \cite[Thm.~4.2]{Naik22}
and Bayesian linear regression \cite[Prop.~3]{Jankowiak22}.

\cref{cor:subopt} generalizes these results substantially, and demonstrates that coresets of size $M = O(\polylog(N))$ produced by
the subsample-optimize method in \cref{alg:opt} maintain a bounded KL divergence as $N\to\infty$.
Two key assumptions are subexponentiality of the potentials and a polynomial (in $N$) growth of $\Var_\pi(\ell(\theta))$;
these conditions are not stringent and should hold for a wide range of Bayesian models and \iid data generating processes. 
The last key assumption in \cref{eq:alignmentpr} is that a randomly-chosen potential function $\ell_I$, $I\dist\Cat(p_1,\dots,p_N)$
(with probabilities as in \cref{alg:opt}) is well-aligned with the residual coreset error function. Similar alignment
conditions have appeared in past results for more restrictive settings (see, e.g., $J(\delta)$ in \citep[Thm.~4.1]{Naik22}).

\bcor\label{cor:subopt}
Suppose there exist $\beta,\alpha > 0$ and $0\leq \rho,\eps < 1$ such that 
the potential functions $(\ell_n)_{n=1}^N$ are $\beta\Cov_\pi((\ell_n)_{n=1}^N)$-subexponential with probability increasing to 1 as $N\to\infty$, 
$\Var_\pi(\ell(\theta)) = O_p(N^\alpha)$, $M = (\log N)^{\frac{1}{1-\rho}}$, and 
\[
\P\lt(\max\lt\{0, \Corr_\pi\lt(\ell_{I_M}(\theta), \ell(\theta) - \ell_{M-1}^\star(\theta)\rt)\rt\}^2 \geq 1-\eps \m| (\ell_n)_{n=1}^N\rt) = \omega_p(M^{-\rho})\label{eq:alignmentpr}\\
\ell^\star_{M-1}(\theta) = \hspace{-.5cm}\argmin_{g \in \cone\{\ell_{I_1},\dots,\ell_{I_{M-1}}\}} \hspace{-.5cm}\Var_\pi\lt(\ell(\theta)-g(\theta)\rt) \qquad I_1,\dots, I_M\distiid\Cat(p_1,\dots,p_N).
\]
Then \cref{alg:opt} produces a coreset with $\KLmax(w) = O_p(1)$ as $N\to\infty$.
\ecor

\begin{algorithm}[t]
\caption{Subsample-optimize coreset construction}\label{alg:opt}
\begin{algorithmic}
\State Compute probabilities $(p_n)_{n=1}^N$ (may depend on the data and model)
\State Draw $I_1, \dots, I_M \distiid \Cat(p_1,\dots,p_N)$, and set $\scI = \{I_1, \dots, I_M\}$
\State Compute $w^\star = \argmin_{w\in\reals_+^N}\KL(\pi_w||\pi) \quad \text{s.t.} \quad w_n \neq 0 \text{ only if } n \in \scI.$
\State\Return $(w^\star_n)_{n=1}^N$
\end{algorithmic}
\end{algorithm}

\bfig
\centering\includegraphics[width=0.49\textwidth]{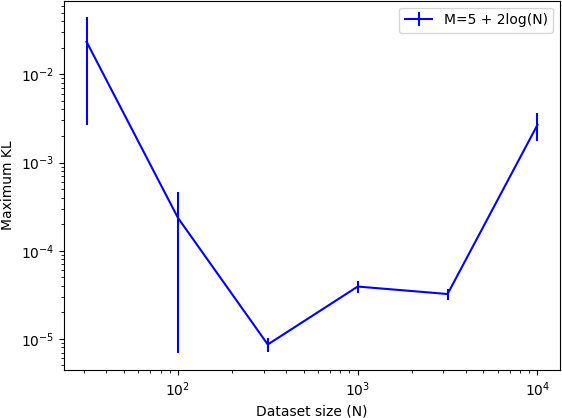}
\hfill
\centering\includegraphics[width=0.49\textwidth]{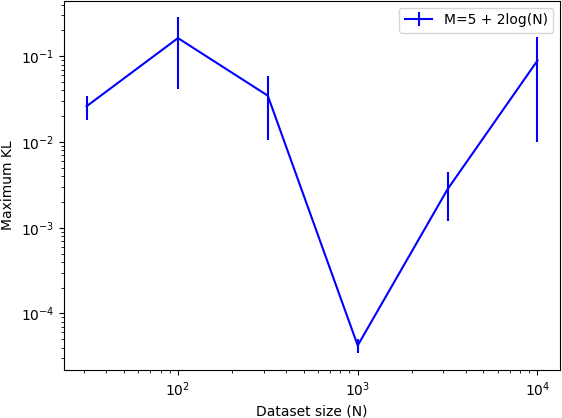}
\caption{Subsample-optimize coreset quality, showing the maximum of the forward and reverse KL divergences on the vertical axis
as a function of dataset size $N$ for coresets of size $5+2\log N$. 
Solid lines indicate the mean over 70 trials, and error bars indicate standard error.
The left panel is for the Cauchy location model,
while the right panel is for the logistic regression model. Sampling probabilities are uniform $p_n=1/N$,
and coreset weights were optimized by nonnegative least squares for log-likelihoods discretized via samples from $\pi$ \cite[Eq.~4]{Campbell19b}.}\label{fig:subopt}
\efig

\cref{fig:subopt} confirms that subsample-optimize coreset construction methods applied to the logistic regression and Cauchy location models
in \cref{eq:cauchymodel,eq:logregmodel} (which both violate the conditions of past upper bounds in the literature)
are able to provide high-quality posterior approximations for very small coresets---in this case, $M \propto \log N$. 

\section{Conclusions}\label{sec:conclusions}
This article presented new general lower and upper bounds on the quality
of Bayesian coreset approximations, as measured by the KL divergence.
These results were used to draw novel conclusions regarding importance-weighted and subsample-optimize coreset methods,
which align with simulation experiments on two synthetic models that violate the assumptions of past theoretical results. 
Avenues for future work include general bounds on the subexponentiality constant
$\beta$ in \cref{prop:allsubexp}, as well as the alignment probability in \cref{eq:alignmentpr}, 
in the setting of Bayesian models with \iid data generating processes. A limitation of this work is that both quantities currently require
case-by-case analysis.

\begin{ack}
The author gratefully acknowledges the support of an NSERC Discovery Grant (RGPIN-2019-03962).
\end{ack}

\small
\bibliographystyle{unsrt}
\bibliography{sources}

\begin{thebibliography}{10}

\bibitem{Robert04}
Christian Robert and George Casella.
\newblock {\em Monte Carlo Statistical Methods}.
\newblock Springer, 2nd edition, 2004.

\bibitem{Gelman13}
Andrew Gelman, John Carlin, Hal Stern, David Dunson, Aki Vehtari, and Donald
  Rubin.
\newblock {\em Bayesian data analysis}.
\newblock CRC Press, 3rd edition, 2013.

\bibitem{Christen05}
J.~Andrés Christen and Colin Fox.
\newblock {M}arkov chain {M}onte {C}arlo using an approximation.
\newblock {\em Journal of Computational and Graphical Statistics},
  14(4):795--810, 2005.

\bibitem{Banterle19}
Marco Banterle, Clara Grazian, Anthony Lee, and Christian~P. Robert.
\newblock Accelerating {M}etropolis-{H}astings algorithms by delayed
  acceptance.
\newblock {\em Foundations of Data Science}, 1(2):103--128, 2019.

\bibitem{Payne14}
Richard Payne and Bani Mallick.
\newblock {B}ayesian big data classification: a review with complements.
\newblock {\em arXiv:1411.5653}, 2014.

\bibitem{Sherlock17}
Chris Sherlock, Andrew Golightly, and Daniel Henderson.
\newblock Adaptive, delayed-acceptance {MCMC} for targets with expensive
  likelihoods.
\newblock {\em Journal of Computational and Graphical Statistics},
  26(2):434--444, 2017.

\bibitem{Doucet15}
Arnaud Doucet, Michael Pitt, George Deligiannidis, and Robert Kohn.
\newblock Efficient implementation of {M}arkov chain {M}onte {C}arlo when using
  an unbiased likelihood estimator.
\newblock {\em Biometrika}, 102(2):295--313, 2015.

\bibitem{Maclaurin14}
Dougal Maclaurin and Ryan Adams.
\newblock Firefly {M}onte {C}arlo: exact {MCMC} with subsets of data.
\newblock In {\em Conference on Uncertainty in Artificial Intelligence}, 2014.

\bibitem{Quiroz21}
Matias Quiroz, Minh-Ngoc Tran, Mattias Villani, Robert Kohn, and Khue-Dung
  Dang.
\newblock The block-{P}oisson estimator for optimally tuned exact subsampling
  {MCMC}.
\newblock {\em Journal of Computational and Graphical Statistics},
  30(4):877--888, 2021.

\bibitem{Welling11}
Max Welling and Yee~Whye Teh.
\newblock Bayesian learning via stochastic gradient {L}angevin dynamics.
\newblock In {\em International Conference on Machine Learning}, 2011.

\bibitem{Ahn12}
Sungjin Ahn, Anoop Korattikara, and Max Welling.
\newblock {B}ayesian posterior sampling via stochastic gradient {F}isher
  scoring.
\newblock In {\em International Conference on Machine Learning}, 2012.

\bibitem{Korattikara14}
Anoop Korattikara, Yutian Chen, and Max Welling.
\newblock Austerity in {MCMC} land: cutting the {M}etropolis-{H}astings budget.
\newblock In {\em International Conference on Machine Learning}, 2014.

\bibitem{Chen15}
Tianqi Chen, Emily Fox, and Carlos Guestrin.
\newblock Stochastic gradient {H}amiltonian {M}onte {C}arlo.
\newblock In {\em International Conference on Machine Learning}, 2015.

\bibitem{Johndrow20}
James Johndrow, Natesh Pillai, and Aaron Smith.
\newblock No free lunch for approximate {MCMC}.
\newblock {\em arXiv:2010.12514}, 2020.

\bibitem{Bardenet17}
R\'emi Bardenet, Arnaud Doucet, and Chris Holmes.
\newblock On {M}arkov chain {M}onte {C}arlo methods for tall data.
\newblock {\em Journal of Machine Learning Research}, 18:1--43, 2017.

\bibitem{Nagapetyan17}
Tigran Nagapetyan, Andrew Duncan, Leonard Hasenclever, Sebastian Vollmer,
  Lukasz Szpruch, and Konstantinos Zygalakis.
\newblock The true cost of stochastic gradient {L}angevin dynamics.
\newblock {\em arXiv:1706.02692}, 2017.

\bibitem{Baker19}
Jack Baker, Paul Fearnhead, Emily Fox, and Christopher Nemeth.
\newblock Control variates for stochastic gradient {MCMC}.
\newblock {\em Statistics and Computing}, 29:599--615, 2019.

\bibitem{Nemeth21}
Christopher Nemeth and Paul Fearnhead.
\newblock Stochastic gradient {M}arkov {C}hain {M}onte {C}arlo.
\newblock {\em Journal of the American Statistical Association},
  116(533):433--450, 2021.

\bibitem{Quiroz19}
Matias Quiroz, Robert Kohn, Mattias Villani, and Minh-Ngoc Tran.
\newblock Speeding up {MCMC} by efficient data subsampling.
\newblock {\em Journal of the American Statistical Association},
  114(526):831--843, 2019.

\bibitem{Quiroz18}
Matias Quiroz, Robert Kohn, and Khue-Dung Dang.
\newblock Subsampling {MCMC}---an introduction for the survey statistician.
\newblock {\em Sankhya: The Indian Journal of Statistics}, 80-A:S33--S69, 2018.

\bibitem{Blei17}
David Blei, Alp Kucukelbir, and Jon McAuliffe.
\newblock Variational inference: A review for statisticians.
\newblock {\em Journal of the American Statistical Association},
  112(518):859--877, 2017.

\bibitem{Shun95}
Zhenming Shun and Peter McCullagh.
\newblock Laplace approximation of high dimensional integrals.
\newblock {\em Journal of the Royal Statistical Society: Series B},
  57(4):749--760, 1995.

\bibitem{Hall11}
Peter Hall, Tung Pham, Matt Wand, and Shen~S.J. Wang.
\newblock Asymptotic normality and valid inference for gaussian variational
  approximation.
\newblock {\em The Annals of Statistics}, 39(5):2502--2532, 2011.

\bibitem{vanderVaart00}
Aad van~der Vaart.
\newblock {\em Asymptotic Statistics}.
\newblock Cambridge University Press, 2000.

\bibitem{Wang18}
Yixin Wang and David Blei.
\newblock Frequentist consistency of variational {B}ayes.
\newblock {\em Journal of the American Statistical Association},
  114(527):1147--1161, 2018.

\bibitem{CheriefAbdellatif18}
Badr-Eddine Ch\'erief-Abdellatif and Pierre Alquier.
\newblock Consistency of variational {B}ayes inference for estimation and model
  selection in mixtures.
\newblock {\em Electronic Journal of Statistics}, 12:2995--3035, 2018.

\bibitem{Yang18}
Yun Yang, Debdeep Pati, and Anirban Bhattacharya.
\newblock $\alpha$-variational inference with statistical guarantees.
\newblock {\em The Annals of Statistics}, 2018.

\bibitem{Alquier20}
Pierre Alquier and James Ridgway.
\newblock Concentration of tempered posteriors and of their variational
  approximations.
\newblock {\em The Annals of Statistics}, 48(3):1475--1497, 2020.

\bibitem{Xu22}
Zuheng Xu and Trevor Campbell.
\newblock The computational asymptotics of {G}aussian variational inference and
  the {L}aplace approximation.
\newblock {\em Statistics and Computing}, 32(63), 2022.

\bibitem{Miller21}
Jeffrey Miller.
\newblock Asymptotic normality, concentration, and coverage of generalized
  posteriors.
\newblock {\em Journal of Machine Learning Research}, 22:1--53, 2021.

\bibitem{Zhang19}
Cheng Zhang, Judith B\"utepage, Hedvig Kjellstr\"om, and Stephan Mandt.
\newblock Advances in variational inference.
\newblock {\em IEEE Transactions on Pattern Analysis and Machine Intelligence},
  41(8):2008--2026, 2019.

\bibitem{Huggins16}
Jonathan Huggins, Trevor Campbell, and Tamara Broderick.
\newblock Coresets for scalable {B}ayesian logistic regression.
\newblock In {\em Advances in Neural Information Processing Systems}, 2016.

\bibitem{Campbell19JMLR}
Trevor Campbell and Tamara Broderick.
\newblock Automated scalable {B}ayesian inference via {H}ilbert coresets.
\newblock {\em Journal of Machine Learning Research}, 20(15):1--38, 2019.

\bibitem{Campbell19b}
Trevor Campbell and Boyan Beronov.
\newblock Sparse variational inference: {B}ayesian coresets from scratch.
\newblock In {\em Advances in Neural Information Processing Systems}, 2019.

\bibitem{Campbell18}
Trevor Campbell and Tamara Broderick.
\newblock Bayesian coreset construction via greedy iterative geodesic ascent.
\newblock In {\em International Conference on Machine Learning}, 2018.

\bibitem{Chen22}
Naitong Chen, Zuheng Xu, and Trevor Campbell.
\newblock {B}ayesian inference via sparse {H}amiltonian flows.
\newblock In {\em Advances in Neural Information Processing Systems}, 2022.

\bibitem{Naik22}
Cian Naik, Judith Rousseau, and Trevor Campbell.
\newblock Fast {B}ayesian coresets via subsampling and quasi-{N}ewton
  refinement.
\newblock In {\em Advances in Neural Information Processing Systems}, 2022.

\bibitem{Jankowiak22}
Martin Jankowiak and Du~Phan.
\newblock Surrogate likelihoods for variational annealed importance sampling.
\newblock In {\em International Conference on Machine Learning}, 2022.

\bibitem{Chen24}
Naitong Chen and Trevor Campbell.
\newblock Coreset {M}arkov chain {M}onte {C}arlo.
\newblock In {\em International Conference on Artificial Intelligence and
  Statistics}, 2024.

\bibitem{Cormen22}
Thomas Cormen, Charles Leiserson, Ronald Rivest, and Clifford Stein.
\newblock {\em Introduction to Algorithms}.
\newblock The MIT Press, $4^\text{th}$ edition, 2022.

\bibitem{Ma15}
Ping Ma, Michael Mahoney, and Bin Yu.
\newblock A statistical perspective on algorithmic leveraging.
\newblock {\em Journal of Machine Learning Research}, 16:861--911, 2015.

\bibitem{Wang18b}
HaiYing Wang, Rong Zhu, and Ping Ma.
\newblock Optimal subsampling for large sample logistic regression.
\newblock {\em Journal of the American Statistical Association},
  113(522):829--844, 2018.

\bibitem{Wang19}
HaiYing Wang.
\newblock More efficient estimation for logistic regression with optimal
  subsamples.
\newblock {\em Journal of Machine Learning Research}, 20:1--59, 2019.

\bibitem{Ai21}
Mingyao Ai, Jun Yu, Huiming Zhang, and HaiYing Wang.
\newblock Optimal subsampling algorithms for big data regressions.
\newblock {\em Statistica Sinica}, 31(2):749--772, 2021.

\bibitem{Wang21}
HaiYing Wang and Yanyuan Ma.
\newblock Optimal subsampling for quantile regression in big data.
\newblock {\em Biometrika}, 108(1):99--112, 2021.

\bibitem{Feldman20}
Dan Feldman.
\newblock Introduction to core-sets: an updated survey.
\newblock {\em arXiv:2011.09384}, 2020.

\bibitem{Walker74}
Alastair Walker.
\newblock New fast method for generating discrete random numbers with arbitrary
  frequency distributions.
\newblock {\em Electronics Letters}, 10(8):127--128, 1974.

\bibitem{Walker77}
Alastair Walker.
\newblock An efficient method for generating discrete random variables with
  general distributions.
\newblock {\em ACM Transactions on Mathematical Software}, 3(3):253--256, 1977.

\bibitem{vanErven14}
Tim van Erven and Peter Harr{\"e}mos.
\newblock {R}{\'e}nyi divergence and {K}ullback-{L}eibler divergence.
\newblock {\em IEEE Transactions on Information Theory}, 60(7):3797--3820,
  2014.

\bibitem{Vershynin20}
Roman Vershynin.
\newblock {\em High-dimensional probability: an introduction with applications
  in data science}.
\newblock Cambridge University Press, 2020.

\bibitem{Vajda70}
Igor Vajda.
\newblock Note on discrimination information and variation.
\newblock {\em IEEE Transactions on Information Theory}, 16(6):771--773, 1970.

\bibitem{Pollard02}
David Pollard.
\newblock {\em A user's guide to probability theory}.
\newblock Cambridge series in statistical and probabilistic mathematics.
  Cambridge University Press, $7^\text{th}$ edition, 2002.

\bibitem{Keener10}
Robert Keener.
\newblock {\em Theoretical statistics: topics for a core course}.
\newblock Springer, 2010.

\bibitem{Bulinski17}
Andre Bulinski.
\newblock Conditional central limit theorem.
\newblock {\em Theory of Probability \& its Applications}, 61(4):613--631,
  2017.

\bibitem{Schwartz65}
Lorraine Schwartz.
\newblock On {B}ayes procedures.
\newblock {\em Z. Wahrscheinlichkeitstheorie und Verwandte Gebiete}, 4:10--26,
  1965.

\end{thebibliography}

\newpage
\appendix
\section{Proofs}\label{sec:proofs}
\bprfof{\cref{lem:kl_lower}}
By Vajda's inequality \cite{Vajda70},
\[
\KLmin(w)
 &\geq \log\frac{1+\TV(\pi,\pi_w)}{1-\TV(\pi,\pi_w)} - \frac{2\TV(\pi,\pi_w)}{1+\TV(\pi,\pi_w)}\\
 &\geq -\log\left(1-\TV(\pi,\pi_w)\right) - \TV(\pi,\pi_w)\\
 &\geq 0.
\]
The bound is monotone increasing in $\TV(\pi,\pi_w)$; therefore because
the squared Hellinger distance satisfies the inequality \cite[p.~61]{Pollard02},
\[
H^2(\pi,\pi_w) = \frac{1}{2}\int \left(\sqrt{\pi}-\sqrt{\pi_w}\right)^2 \leq \frac{1}{2}\int\left|\pi-\pi_w\right| = \TV(\pi,\pi_w),
\]
we have that 
\[
\KLmin(w)
&\geq -\log\left(1-H^2(\pi,\pi_w)\right) - H^2(\pi,\pi_w).
\]
We substitute the value of the squared Hellinger distance to find that 
\[
\KLmin(w) &\geq -\log\lt(\int \sqrt{\pi\pi_w}\rt) + \int\sqrt{\pi\pi_w}- 1 \geq 0.
\]
Note that $\int\sqrt{\pi\pi_w} \leq 1$, so
\[
\KLmin(w) &\geq -\log\lt(\min\{1,\int \sqrt{\pi\pi_w}\}\rt) + \min\{1,\int\sqrt{\pi\pi_w}\}- 1 \geq 0.
\]
The bound is monotone decreasing in $\int\sqrt{\pi\pi_w}$, so we require an upper bound on $\int\sqrt{\pi\pi_w}$. 
To obtain the required bound,
we split the integral into two parts---one on the set $B$,
and the other on $B^c$---and then use the Cauchy-Schwarz inequality to bound the part on $B^c$. Note that by definition $\pi$ and $\pi_w$ are mutually dominating,
so the density ratio $\pi_w/\pi$ is well-defined and measurable.
\[
\int\sqrt{\pi\pi_w} &= \int_B\sqrt{\pi\pi_w} + \int_{B^c}\sqrt{\pi\pi_w}\\
&= \int_B\sqrt{\pi\pi_w} + \int \pi \sqrt{\frac{\pi_w}{\pi}}\1_{B^c}\\
&\leq \int_B\sqrt{\pi\pi_w} + \sqrt{\pi(B^c)}\\
&= \frac{\int_B\pi_0\exp\frac{1}{2}(\ell + \ell_w)}{\sqrt{\int \pi_0\exp(\ell) \int\pi_0\exp(\ell_w)}} + \sqrt{\pi(B^c)}.
\]
The result follows.
\eprfof

\bprfof{\cref{lem:kl_upper}}
We first consider the forward KL divergence. By definition,
\[
\KL(\pi||\pi_w) &= \int \pi (\ell-\ell_w) + \log\frac{\int \pi_0\exp(\ell_w)}{\int \pi_0\exp(\ell)}\\
&= \int \pi (\ell-\ell_w) + \log\int \pi \exp(\ell_w-\ell).
\]
Since the $\KL$ is positive, for $\lambda > 0$,
\[
\KL(\pi||\pi_w) &\leq \frac{1+\lambda}{\lambda}\int \pi (\ell-\ell_w) + \frac{1+\lambda}{\lambda}\log\int \pi \exp(\ell_w-\ell)\\
&\leq \frac{1+\lambda}{\lambda}\int \pi (\ell-\ell_w) + \frac{1}{\lambda}\log\int \pi \exp((1+\lambda)(\ell_w-\ell))\\
&= \frac{1}{\lambda}\log\int \pi \exp((1+\lambda)(\bar\ell_w-\bar\ell)),
\]
by Jensen's inequality.
Next we consider the reverse KL divergence. For any $\lambda \neq 0$,
\[
\KL(\pi_w||\pi) &= \int \pi_w(\ell_w-\ell) + \log \frac{\int \pi_0\exp(\ell)}{\int \pi_0\exp(\ell_w)}\\
&= \frac{1}{\lambda} \int \pi_w \lambda(\ell_w-\ell) - \log \int \pi \exp(\ell_w-\ell).
\]
By Jensen's inequality, for $\lambda > 0$,
\[
\KL(\pi_w||\pi) &\leq \frac{1}{\lambda}\log \int \pi_w \exp(\lambda(\ell_w-\ell)) - \log \int \pi \exp(\ell_w-\ell)\\
&= \frac{1}{\lambda}\log \frac{\int \pi\exp((1+\lambda)(\ell_w-\ell))}{\int \pi\exp(\ell_w-\ell)} - \log \int \pi \exp(\ell_w-\ell)\\
&= \frac{1}{\lambda}\log \int \pi\exp((1+\lambda)(\ell_w-\ell)) -\frac{1+\lambda}{\lambda} \log \int \pi \exp(\ell_w-\ell)\\
&\leq \frac{1+\lambda}{\lambda} \int \pi (\ell-\ell_w) + \frac{1}{\lambda}\log \int \pi\exp((1+\lambda)(\ell_w-\ell))\\
&= \frac{1}{\lambda}\log\int \pi \exp((1+\lambda)(\bar\ell_w-\bar\ell)).
\]
This is the same bound as in the forward KL divergence case.
Since the bound applies for all $\lambda > 0$, we can take the infimum.
\eprfof

\bprfof{\cref{thm:diffable_lower}}
By replacing the integrals over the whole space $\Theta$ in the denominator of $J_B(w)$ in \cref{lem:kl_lower} with integrals over the subset $B$, 
\[
\KLmin(w) &\geq -\log \min(1,J_B(w)) + \min(1,J_B(w)) - 1\\
&\geq O_p(1) -\log J_B(w)\\
&\geq O_p(1)+ \min\lt\{G_B(w), -\log\sqrt{\pi(B^c)}\rt\}\\
G_B(w) &= -\log \int_B \pi_0 \exp((1/2)(\ell+\ell_w)) + \frac{1}{2}\log\int_B\pi_0\exp(\ell) + \frac{1}{2}\log\int_B\pi_0\exp(\ell_w).
\]
So to obtain the stated lower bound on the KL divergence, we require 
an upper bound on $\log\int_B \pi_0 \exp((1/2)(\ell+\ell_w))$, 
and lower bounds on $\log\int_B \pi_0\exp(\ell)$ and $\log\int_B\pi_0\exp(\ell_w)$.
By Taylor's theorem, \cref{assum:lower}, and the assumption on $\grad^2\ell_w(\theta)$,
for all $\theta \in B$,
\[
\begin{aligned}
\lt| \ell(\theta) - \ell(\theta_0)- g^T(\theta-\theta_0) + \frac{N}{2}(\theta-\theta_0)^TH(\theta-\theta_0)\rt|
&\leq \frac{N o_p(1) }{2}(\theta-\theta_0)^TH(\theta-\theta_0)\\
\lt| \ell_w(\theta) - \ell_w(\theta_0) - g_w^T(\theta-\theta_0) + \frac{\sbw}{2}(\theta-\theta_0)^TH(\theta-\theta_0)\rt|
&\leq  \frac{\sbw o_p(1)}{2}(\theta-\theta_0)^TH(\theta-\theta_0).
\end{aligned}\label{eq:taylorexpansions}
\]
We shift the exponential arguments in $G_B(w)$ by $(1/2)(\ell(\theta_0)+\ell_w(\theta_0))$,
note that $\pi_0$ is continuous and positive around $\theta_0$, and
and apply the Taylor expansions in \cref{eq:taylorexpansions} to obtain an upper bound on the first term:
\[
\log\int_B \pi_0 e^{\frac{1}{2}(\ell-\ell(\theta_0) + \ell_w-\ell_w(\theta_0))}
&\leq O_p(1) + \log \int_B e^{\frac{1}{2}((g+g_w)^T(\theta-\theta_0) - \frac{\oneish(N+\sbw)}{4}(\theta-\theta_0)^TH(\theta-\theta_0)},
\]
where $\oneish$ denotes a quantity that converges in probability to 1 as $N\to\infty$.
We can transform variables to $x = C^T(\theta-\theta_0)$, where $H=CC^T$ is the Cholesky factorization of $H$,
and subsequently complete the square:
\[
\log\int_B \pi_0 e^{\frac{1}{2}(\dots)}
&\leq O_p(1) +\frac{\oneish\|C^{-1}(g+g_w)\|^2}{4(N+\sbw)} + \log \int_{\|x\|^2\leq r^2} e^{- \frac{\oneish(N+\sbw)}{4}\lt\|x - \frac{\oneish C^{-1}(g+g_w)}{(N+\sbw)}\rt\|^2}.\label{eq:upperproto}
\]
We can obtain lower bounds on the other two terms using a similar technique:
\[
\log \int_B \pi_0 e^{\ell-\ell(\theta_0)} 
&\geq O_p(1)  + \frac{\oneish\|C^{-1}g\|^2}{2N} + \log \int_{\|x\|^2\leq r^2} e^{-\frac{\oneish N}{2} \lt\|x - \frac{\oneish C^{-1}g}{N}\rt\|^2}\label{eq:lowerproto}\\
\log \int_B \pi_0 e^{\ell_w-\ell_w(\theta_0)}
&\geq O_p(1)  + \frac{\oneish\|C^{-1}g_w\|^2}{2\sbw} + \log \int_{\|x\|^2\leq r^2} e^{-\frac{\oneish\sbw}{2} \lt\|x - \frac{\oneish C^{-1}g_w}{\sbw}\rt\|^2}.\label{eq:lowerprotow}
\]
It remains to analyze the three $\log \int \dots$ terms. We bound the integral term in \cref{eq:upperproto} with the integral over the whole space:
\[
\log\int_{\|x\|^2\leq r^2} e^{-\frac{\oneish(N+\sbw)}{4}\|\dots\|^2} &\leq O_p(1) - \frac{d}{2}\log \lt(N+\sbw\rt).
\]
For the integral term in \cref{eq:lowerproto}, note that since $Nr^2 = \omega(1)$ and $\|C^{-1}g/N\| = O_p(N^{-1/2})$, we have
\[
&\log\int_{\|x\|^2 \leq r^2} e^{-\frac{\oneish N}{2}\|\dots\|^2} \\
&= \log\lt(\int e^{-\frac{\oneish N}{2}(\dots)} - \int_{\|x\|^2 > r^2} e^{-\frac{\oneish N}{2}\|x-\frac{\oneish C^{-1}g}{N}\|^2}\rt)\\
&\geq \log\lt(\lt(\frac{\oneish 2\pi}{N}\rt)^{d/2} - e^{-\frac{\oneish N}{4}\min_{\|x\|\geq r}\lt\|x - \frac{\oneish C^{-1}g}{N}\rt\|^2} \int e^{-\frac{\oneish N}{4}\|x-\frac{\oneish C^{-1}g}{N}\|^2}\rt)\\
&= \log\lt(\lt(\frac{\oneish 2\pi}{N}\rt)^{d/2} - e^{-\frac{\Omega_p(Nr^2)}{4}} \lt(\frac{\oneish 4\pi}{N}\rt)^{d/2}\rt)\\
&= -\frac{d}{2}\log(N) + O_p(1).
\]
For the integral term in \cref{eq:lowerprotow}, we consider two cases: one where $\sbw$ is large, and one where it is small.
First assume $\sbw r^2 > 8d\log 2$; then by a similar technique as used in the first lower bound,
since $\|C^{-1}g_w/\sbw\| = o_p(r)$, 
\[
&\log\int_{\|x\|^2 \leq r^2} e^{-\frac{\oneish \sbw}{2}\|\dots\|^2} \\
&\geq \log\lt(\lt(\frac{\oneish 2\pi}{\sbw}\rt)^{d/2} - e^{-\frac{\oneish \sbw}{4}\min_{\|x\|\geq r} \lt\|x - \frac{\oneish C^{-1}g_w}{\sbw}\rt\|^2} \int e^{-\frac{\oneish \sbw}{4}\|x-\frac{\oneish C^{-1}g_w}{\sbw}\|^2}\rt)\\
&\geq \log\lt(\lt(\frac{\oneish 2\pi}{\sbw}\rt)^{d/2} - e^{-2d\log 2\oneish}\lt(\frac{\oneish 4\pi}{\sbw}\rt)^{d/2}\rt)\\
&\geq - \frac{d}{2}\log\sbw  + O_p(1).
\]
When $\sbw r^2 \leq 8d\log 2$, we transform variables $y = x/r$ to find that
since $\|C^{-1}g_w/\sbw\| = o_p(r)$,
\[
\log \int_{\|x\|^2\leq r^2} e^{-\frac{\oneish \sbw}{2}\|\dots\|^2}
&= \frac{d}{2}\log r^2 + \log\int_{\|y\|^2\leq 1} e^{-\frac{\oneish \sbw r^2}{2} \lt\|y - \frac{\oneish C^{-1}g_w}{r\sbw}\rt\|^2}\\
&\geq \frac{d}{2}\log r^2 + \log e^{-\frac{8d\log 2 \oneish}{2}\lt(2 + 2\lt\|\frac{\oneish C^{-1}g_w}{r\sbw}\rt\|^2\rt)} \lt(\int_{\|y\|^2\leq 1} 1\rt)\\
&= \frac{d}{2}\log r^2  + O_p(1).
\]
Therefore regardless of the value of $\sbw$,
\[
\log \int_{\|x\|^2\leq r^2} e^{-\frac{\oneish \sbw}{2}\|\dots\|^2} &\geq -\frac{d}{2}\log\lt(\max\{\sbw, 1/r^2\}\rt) + O_p(1).
\]
So therefore combining all previous results,
\[
G_B(w) &\geq 
O_p(1) 
+ \frac{\oneish}{4}\lt( \frac{\|C^{-1}g\|^2}{N} + \frac{\|C^{-1}g_w\|^2}{\sbw} - \frac{\|C^{-1}(g+g_w)\|^2}{N+\sbw}\rt)
+ \frac{d}{4}\log\frac{(N+\sbw)^2}{N\max\{\sbw, 1/r^2\}}\\
&= 
O_p(1) 
+\frac{\oneish}{4}\lt( \frac{\sbw\|C^{-1}g\|^2}{N(N+\sbw)} + \frac{N\|C^{-1}g_w\|^2}{\sbw(N+\sbw)} - \frac{2g^TH^{-1}g_w}{N+\sbw}\rt)
+ \frac{d}{4}\log\frac{(N+\sbw)^2}{N\max\{\sbw, 1/r^2\}}\\
&=
O_p(1) 
+\frac{\oneish}{4}\lt(\frac{N\sbw}{N+\sbw}\lt\|\frac{C^{-1}g}{N} - \frac{C^{-1}g_w}{\sbw}\rt\|^2\rt)
+ \frac{d}{4}\log\frac{(N+\sbw)^2}{N\max\{\sbw, 1/r^2\}}\\
&=
O_p(1) 
+\Omega_p(1)\lt(
\frac{N\sbw}{N+\sbw}\lt\|\frac{g}{N} - \frac{g_w}{\sbw}\rt\|^2
+ d\log\frac{(N+\sbw)^2}{N\max\{\sbw, 1/r^2\}}\rt).
\]
We now consider the minimum over $\alpha \geq 0$. Since neither $O_p(1)$ or $\Omega_p(1)$ above depends on $\sbw$, we have that
\[
\min_{\alpha \geq 0} \KLmin(\alpha w) &\geq O_p(1) + \Omega_p(1)
\min\lt\{-\log \pi(B^c), \lt(\min_{\alpha \geq 0}\frac{ N\alpha\sbw}{N+\alpha \sbw}\lt\|\frac{g}{N}-\frac{g_w}{\sbw}\rt\|^2 + d\log\frac{(N+\alpha \sbw)^2}{N\max\{\alpha \sbw,1/r^2\}}\rt)\rt\}.
\]
On the $1/r^2$ branch of the objective function, the derivative in $\alpha$ is always positive, and hence the minimum occurs at $\alpha = 0$, and so
\[
\min_{\alpha \geq 0} (\dots) &\geq d\log(Nr^2).
\]
On the $\alpha w$ branch of the objective function,
\[
\min_{\alpha \geq 0}\frac{ N\alpha\sbw}{N+\alpha \sbw}\lt\|\frac{g}{N}-\frac{g_w}{\sbw}\rt\|^2 + d\log\frac{(N+\alpha \sbw)^2}{N\alpha \sbw}
&\geq
\min_{\alpha \geq 0}\frac{ N\alpha\sbw}{N+\alpha \sbw}\lt\|\frac{g}{N}-\frac{g_w}{\sbw}\rt\|^2 + d\log\frac{(N+\alpha \sbw)}{N\alpha \sbw} + d\log N.
\]
For $a, b > 0$ and $x \geq 0$, the function $a x - b \log x$ is convex in $x$ with minimum at $x^\star = b/a$, and so
\[
\min_{\alpha \geq 0} (\dots)
&\geq d\log\lt(N\lt\|\frac{g}{N}-\frac{g_w}{\sbw}\rt\|^2\rt).
\]
By assumption, $\|\frac{g}{N}\| = o_p(r)$ and $\|\frac{g_w}{\sbw}\| = o_p(r)$, and hence the $\alpha w$ branch has the asymptotic minimum:
\[
\min_{\alpha \geq 0} \KLmin(\alpha w) &\geq O_p(1) + \Omega_p(1)\min\lt\{-\log\pi(B^c), d\log\lt(N\lt\|\frac{g}{N}-\frac{g_w}{\sbw}\rt\|^2\rt)\rt\}.
\]
\eprfof

\bprfof{\cref{thm:diffable_lower_min}}
By \cref{lem:kl_lower},
\[
\KLmin(w) &\geq -\log \min(1,J_B(w)) + \min(1,J_B(w)) - 1\\
&\geq O_p(1) + \min\lt\{ G_B(w), -\log\sqrt{\pi(B^c)}\rt\}\\
G_B(w) &= -\log \int_B \pi_0 \exp((1/2)(\ell+\ell_w)) + \frac{1}{2}\log\int\pi_0\exp(\ell) + \frac{1}{2}\log\int\pi_0\exp(\ell_w).
\]
Note that $G_B$ in this proof is subtly different from the $G_B$ used in the proof of \cref{thm:diffable_lower}; the latter
two integrals are over the whole space (directly from \cref{lem:kl_lower}), rather than $B$.
We shift the exponential arguments in $G_B(w)$ by $(1/2)(\ell(\theta_0)+\ell_w(\theta_0))$.
We first provide lower bounds on two of the integral terms via \cref{assum:lower2}:
\[
\log\int\pi_0e^{\ell-\ell(\theta_0)} &\geq 
O_p(1) + \log\int e^{(g+g_0)^T(\theta-\theta_0) - \frac{\oneish(N+1)L'^2}{2}\|\theta-\theta_0\|^2},
\]
where $\oneish$ denotes a quantity that converges in probability to 1,
$g_0 = \grad\log\pi_0(\theta_0)$, and $L'^2 = \frac{NL^2+L_0^2}{N+1}$.
Transforming variables via $x = L'(\theta-\theta_0)$,
\[
\log\int\pi_0e^{\ell-\ell(\theta_0)} &\geq O_p(1) + \log\int e^{(g+g_0)^Tx/L' - \frac{\oneish (N+1)}{2}\|x\|^2}\\
&= O_p(1) + \log\int e^{-\frac{\oneish (N+1)}{2}\lt\|x - \frac{g+g_0}{(N+1)L'}\rt\|^2 + \frac{\oneish (N+1)}{2}\|\frac{g+g_0}{(N+1)L'}\|^2}\\
&= O_p(1) + \frac{\oneish(N+1)}{2L'^2}\lt\|\frac{g+g_0}{N+1}\rt\|^2 - \frac{d}{2}\log(N+1)\\
&\geq O_p(1) + \frac{\oneish(N+1)}{2\max\{L^2, L_0^2\}}\lt\|\frac{g+g_0}{N+1}\rt\|^2 - \frac{d}{2}\log(N+1).
\]
Let $L^2_w = \frac{1}{\sbw}\sum_n w_n L^2_n$. Using the same technique, with $L'^2_w = \frac{\sbw L_w^2+ L_0^2}{\sbw+1}$ and $x = L'_w(\theta-\theta_0)$, 
\[
\log\int\pi_0e^{\ell_w-\ell_w(\theta_0)} &\geq \log\int e^{(g_w+g_0)^T(\theta-\theta_0) - \frac{\oneish(\sbw+1)}{2} L'^2_w \|\theta-\theta_0\|^2}\\
&\geq O_p(1)+  \frac{\sbw+1}{2L'^2_w}\lt\|\frac{g_w+g_0}{\sbw+1}\rt\|^2 + \log\int e^{-\frac{(\sbw+1)}{2}\lt\|x - \frac{g_w+g_0}{(\sbw+1) L'_w}\rt\|^2 }\\
&\geq O_p(1)+  \frac{\sbw+1}{2L'^2_w}\lt\|\frac{g_w+g_0}{\sbw+1}\rt\|^2 + \log\int_{\|x-\frac{g_w+g_0}{(\sbw+1) L'_w}\| \leq (\sbw+1)^{-1/3}} e^{-\frac{\sbw+1}{2}\lt\|x - \frac{g_w+g_0}{(\sbw+1) L'_w}\rt\|^2 }\\
&= O_p(1) + \frac{\sbw+1}{2L'^2_w}\lt\|\frac{g_w+g_0}{\sbw+1}\rt\|^2 -\frac{d}{2}\log(\sbw+1)\\
&\geq O_p(1) + \frac{\sbw+1}{2\max\{\beta L^2, L_0^2\}}\lt\|\frac{g_w+g_0}{\sbw+1}\rt\|^2 -\frac{d}{2}\log(\sbw+1).
\]
For the upper bound on the first term, we use a local quadratic expansion around $\theta_0$,
where $H_0 = -\grad^2\log\pi_0(\theta_0)$, 
\[
\log\int_B \pi_0 e^{\frac{1}{2}(\ell-\ell(\theta_0) + \ell_w-\ell_w(\theta_0))}
&\leq O_p(1) + \log \int_B e^{\frac{1}{2}((g+g_w+2g_0)^T(\theta-\theta_0) - \frac{\oneish(N+\sbw+2)}{4}(\theta-\theta_0)^T\lt(\frac{(N +\alpha \sbw) H +2H_0}{N+\sbw+2}\rt)(\theta-\theta_0)}.
\]
Because $H \succ 0$, we have $(N+\alpha \sbw)H + 2H_0 \succ 0$ eventually;
we can transform variables to $x = C^T(\theta-\theta_0)$, where $\frac{(N+\alpha\sbw)H+2H_0}{N+\sbw+2}=CC^T$ is the Cholesky factorization,
and subsequently complete the square. Note that
\[
\sqrt{\min\{ \min(\alpha,1) \lambda_{\min} H, \lambda_{\min} H_0\}} \leq \lambda_{\min}C \leq \lambda_{\max} C \leq \sqrt{\max\{ \max(\alpha,1) \lambda_{\max} H, \lambda_{\max} H_0\} }
\] 
so
\[
\log |C| = O_p(1) \qquad \lambda_{\min} C^{-1}HC^{-T} \geq \frac{\lambda_{\min}H}{\max\{\max(\alpha,1) \lambda_{\max}H, \lambda_{\max} H_0\}} = \eta > 0,
\]
and therefore
\[
&\log\int_B \pi_0 e^{\frac{1}{2}(\dots)}\\
&\leq O_p(1) + \frac{\oneish(N+\sbw+2)}{4} \lt\|\frac{C^{-1}(g+g_w+2g_0)}{N+\sbw+2}\rt\|^2 + \log \int_{\|x\|^2\leq r^2\eta^{-1}} e^{-\frac{\oneish(N+\sbw+2)}{4}\lt\|x - \frac{\oneish C^{-1}(g+g_w+2g_0)}{N+\sbw+2}\rt\|^2}.\label{eq:upperint}
\]
Suppose first that $\sbw +1 \leq N/(4\|C^{-1}\|^2\max\{\beta L^2,L_0^2\})$.
In this case we bound the integral in \cref{eq:upperint} by integrating over the whole space:
\[
\log\int_B \pi_0 e^{\frac{1}{2}(\dots)}\leq
O_p(1) + \frac{\oneish\|C^{-1}\|^2(N+\sbw+2)}{4} \lt\|\frac{g+g_w+2g_0}{N+\sbw+2}\rt\|^2 -\frac{d}{2} \log (N+\sbw+2).
\]
Combining this with the previous results yields
\[
&G_B(w) \geq O_p(1)\\
&-\frac{\oneish(N+\sbw+2)}{4}\|C^{-1}\|^2 \lt\|\frac{g+g_w+2g_0}{N+\sbw+2}\rt\|^2 \\
&+\frac{d}{4}\log\frac{(N+\sbw+2)^2}{(N+1)(\sbw+1)}
+ \frac{\sbw+1}{4\max\{\beta L^2,L_0^2\}}\lt\|\frac{g_w+g_0}{\sbw+1}\rt\|^2 +\frac{(N+1)}{4\max\{L^2,L_0^2\}}\lt\|\frac{g+g_0}{N+1}\rt\|^2\\
&\geq O_p(1)+\frac{d}{4}\log\frac{(N+\sbw+2)^2}{(N+1)(\sbw+1)}+\frac{\sbw+1}{4}\lt\|\frac{g_w+g_0}{\sbw+1}\rt\|^2\lt(\frac{1}{\max\{\beta L^2,L_0^2\}} - \frac{2\|C^{-1}\|^2(\sbw+1)}{N+\sbw+2}\rt)\\
&\geq O_p(1)+\frac{d}{4}\log\frac{(N+\sbw+2)^2}{(N+1)(\sbw+1)}+\frac{\sbw+1}{8\max\{\beta L^2,L_0^2\}}\lt\|\frac{g_w+g_0}{\sbw+1}\rt\|^2.
\]
Bounding the last term below by 0 and minimizing over $w$ such that $\sbw \leq \sqrt{N}$ yields
\[
G_B(w) &\geq O_p(1)+\frac{d}{4}\log\sqrt{N} = O_p(1) + \frac{d}{8}\log N.
\]
Bounding $(N+\sbw+2)/(N+1) \geq 1$ and minimizing over $w$ such that $\sbw \geq \sqrt{N}$ yields
\[
G_B(w) &\geq O_p(1)+\frac{d}{4}\log N - \frac{d}{4} \log(\sbw+1)+\frac{\sbw+1}{8\max\{\beta L^2,L_0^2\}}\lt\|\frac{g_w+g_0}{\sbw+1}\rt\|^2\\
&\geq O_p(1)+\frac{d}{4}\log N\lt\|\frac{g_w+g_0}{\sbw+1}\rt\|^2\\
&= O_p(1)+\frac{d}{4}\log N\lt\|\frac{g_w}{\sbw}\rt\|^2,
\]
where the second line follows because 
for $a, b > 0$ and $x \geq 0$, the function $a x - b \log x$ is convex in $x$ with minimum at $x^\star = b/a$.
Therefore for $\sbw+1 \leq N/(\dots)$,
\[
\KLmin(w) &\geq O_p(1) +  \Omega_p(1)d\log\lt(N \min\lt\{\lt\|\frac{g_w}{\sbw}\rt\|^2, 1\rt\}\rt).
\]
Next suppose $\sbw+1 \geq N/(4\|C^{-1}\|^2\max\{\beta L^2,L_0^2\})$.
A second upper bound on \cref{eq:upperint} can be obtained by taking the maximum of the integrand over the integration region $\|x\|^2 \leq r^2$.
Note that since $\|g_w/\sbw\| = \omega_p(r)$, $\sbw = \Omega_p(N)$, $g/N = O_p(N^{-1/2})$, and $Nr^2 = \omega_p(1)$, 
we have that $\|(g+g_w+2g_0)/(N+\sbw+2)\| = \omega_p(r)$, and so
\[
&\log\int_B \pi_0 e^{\frac{1}{2}(\dots)}\\
&\leq O_p(1) + \frac{\oneish(N+\sbw+2)}{4} \lt\|\frac{C^{-1}(g+g_w+2g_0)}{N+\sbw+2}\rt\|^2 
- \frac{\oneish(N+\sbw+2)}{4}\lt(\lt\|\frac{C^{-1}(g+g_w+2g_0)}{N+\sbw+2}\rt\|-r\rt)^2
+\frac{d}{2}\log r^2\\
&= O_p(1) - \frac{\oneish(N+\sbw+2)}{4}r^2 
+ \frac{\oneish(N+\sbw+2)r}{2}\lt\|\frac{C^{-1}(g+g_w+2g_0)}{N+\sbw+2}\rt\|
+\frac{d}{2}\log r^2 .
\]
So therefore combining this result with the previous bounds and minimizing over $\sbw$ yields
\[
G_B(w) &\geq O_p(1)
+ \frac{\oneish(N+\sbw+2)}{4}r^2 
- \frac{\oneish(N+\sbw+2)r}{2}\lt\|\frac{C^{-1}(g+g_w+2g_0)}{N+\sbw+2}\rt\|\\
&-\frac{d}{4}\log((N+1)(\sbw+1)r^4)
+ \frac{\sbw+1}{4\max\{\beta L^2,L_0^2\}}\lt\|\frac{g_w+g_0}{\sbw+1}\rt\|^2 +\frac{(N+1)}{4\max\{L^2,L_0^2\}}\lt\|\frac{g+g_0}{N+1}\rt\|^2\\
&\geq O_p(1)
-\frac{d}{4}\log(Nr^2) + \frac{\oneish N}{4} \lt(\lt\|\frac{g}{N}\rt\| - r\rt)^2 
- \frac{d}{4}\log(\sbw r^2) + \frac{\oneish \sbw}{4}\lt(\lt\|\frac{g_w}{\sbw}\rt\| - r\rt)^2\\
&\geq O_p(1)
-\frac{d}{4}\log(Nr^2) + \frac{\oneish}{4} Nr^2
- \frac{d}{4}\log(r^2) +\frac{d}{4}\log \lt\|\frac{g_w}{\sbw}\rt\|^2\\
&\geq O_p(1)
+\frac{d}{4}\log N\lt\|\frac{g_w}{\sbw}\rt\|^2.
\]
Combining with the earlier bound and noting that $N\min\{\|g_w/\sbw\|, 1\} = \omega_p(1)$ yields the final result.
\eprfof

\bprfof{\cref{cor:lower_other}}
The proof follows directly from \cref{thm:diffable_lower,thm:diffable_lower_min} by the data processing inequality applied to $\KLmin(w)$.
\eprfof

\bprfof{\cref{thm:subexpupper}}
By \cref{lem:kl_upper},
\[
\KLmax(w) &\leq \inf_{\lambda > 0} \frac{1}{\lambda}\log \int \pi \exp\lt((1+\lambda)(\bar\ell_w-\bar\ell)\rt)\\
&=\inf_{\lambda > 0} \frac{1}{\lambda}\log \int \pi \exp\lt(\bar\ell_{(1+\lambda)(w-1)}\rt).
\]
Since $(\ell_n)_{n=1}^N$ are $(f,A)$-subexponential,
if
\[
(1+\lambda)^2(w-1)^T A(w-1) \leq 1, 
\]
then
\[
\int \pi \exp\lt(\bar\ell_{(1+\lambda)(w-1)}\rt) \leq \exp\lt(f((1+\lambda)^2(w-1)^TA(w-1))\rt).
\]
By assumption, the condition holds when $\lambda = 1$; the result follows.
\eprfof

\bprfof{\cref{prop:allsubexp}}
Let $C(w) = \log \int \pi \exp(\bar\ell_w)$.
By the finiteness condition, \cite[Theorem~2.4]{Keener10} asserts that $C(w)$ is continuous, and has derivatives of all orders
that can be obtained by passing differentiation through the integral within the set $\|w\|_2 < \alpha$. 
Let $U = \Cov_\pi((\ell_n)_{n=1}^N)$, and $\scS = \spann\{w \in\reals^N : w^T(\bar\ell_n)_{n=1}^N = 0\,\, \text{$\pi$-a.s.}\}$. Note that
 $\scS = \ker U$: since $w^TUw = \Var_\pi(w^T(\ell_n)_{n=1}^N)$, $w^TUw = 0$ if and only if $w^T(\bar\ell_n)_{n=1}^N = 0$ $\pi$-a.s.;
and since $U$ is symmetric positive semidefinite, $w^TUw = 0$ if and only if $w\in\ker U$.
Therefore $C(w)$ is continuous, has derivatives of all orders, and derivatives can be passed through the integral
within the set $\{w\in\reals^N : w=v+u, \|v\|_2 < \alpha/2, u \in \ker U\}$.
For a vector $w = v+u$, $v\perp \ker U$, $u \in \ker U$,
and minimum positive eigenvalue $\lambda_+$ of $U$,
\[
w^TUw \leq \frac{\alpha^2\lambda_+}{4}  \implies v^TUv \leq \frac{\alpha^2\lambda_+}{4} \implies \|v\|_2 \leq \frac{\alpha}{2},
\]
and so $C(w)$ is continuous, has derivatives of all orders, and derivatives can be passed through the integral
within the set $\{w\in\reals^N : w^TUw \leq \frac{\alpha^2\lambda_+}{4}\}$.
By Taylor's theorem, for any $w$ in this set, there exists 
a distribution $\nu_w$ with density proportional to $\pi\exp(\bar\ell_{w'})$ for some $w'$ on the segment from 
the origin to $w$ such that
\[
C(w) &= \log\int \pi\exp(\bar\ell_w)= \frac{1}{2}w^TUw + \frac{1}{6} \E_{\nu_w}\lt[ (w^T(\bar\ell_n)_{n=1}^N)^3\rt].
\]
By definition of $\nu_w$, 
$w\in\ker U$ implies that $w^T(\bar\ell_n)_{n=1}^N = 0$ $\nu_w$-\as  and hence  $\frac{1}{6} \E_{\nu_w}\lt[ (w^T(\bar\ell_n)_{n=1}^N)^3\rt] = 0$.
Therefore, for $w^TUw \leq \frac{\alpha^2\lambda_+}{4}$,
\[
C(w) &\leq \frac{1}{2}w^TUw\lt(1 + \max_{\begin{subarray}{c}w^TUw \leq \frac{\alpha^2\lambda_+}{4}\\ w\perp \ker U\end{subarray}} \frac{1}{6} \frac{\E_{\nu_w}\lt[ (w^T(\bar\ell_n)_{n=1}^N)^3\rt]}{w^TUw}\rt)\\
&\leq \frac{1}{2}w^TUw\lt(1 + \max_{\|w\|_2 \leq \frac{\alpha}{2}} \frac{1}{6} \frac{\|w\|_2 \lt\|\E_{\nu_w}\lt[(\bar\ell_n)_{n=1}^N\otimes(\bar\ell_n)_{n=1}^N\otimes(\bar\ell_n)_{n=1}^N\rt]\rt\|_2}{\lambda_+}\rt)\\
&\leq \frac{1}{2}w^TUw\lt(1 +  \frac{\alpha}{12\lambda_+} \max_{\|w\|_2 \leq \frac{\alpha}{2}}\lt\|\E_{\nu_w}\lt[(\bar\ell_n)_{n=1}^N\otimes(\bar\ell_n)_{n=1}^N\otimes(\bar\ell_n)_{n=1}^N\rt]\rt\|\rt),
\]
where $\otimes$ denotes outer products to form a tensor. By continuity of derivatives of all orders within the neighbourhood $\|w\|_2 <\alpha$, 
the result follows by selecting a sufficiently small $\alpha$.
\eprfof

\bprop\label{prop:upperrecovered}
Suppose there exist $c\in\reals$, $\alpha,\delta >0$, and $0<\eps<1$ such that $\ell \leq c$ and for all coreset weights $w$ 
satisfying $\alpha w^T\Cov_\pi((\ell_n)_{n=1}^N)w \leq 1$, $|\bar\ell_w| \leq \eps|\ell-c|+\delta$. 
Then
the potentials $(\ell_n)_{n=1}^N$ are $(h,\alpha \Cov_\pi((\ell_n)_{n=1}^N))$-subexponential with 
$h(x) = \frac{1}{2}x + \frac{e^{\delta+c\eps}}{\int \pi_0 e^{\eps\ell}}x^{1-\eps}$.
\eprop

\bprfof{\cref{prop:upperrecovered}}
Let $\ell' = \ell-c$. Since $\ell' \leq 0$ and $|\bar\ell_w| \leq \eps|\ell'| + \delta$ for some $\eps < 1$, $\delta > 0$,
\[
\int \pi\exp(\bar\ell_w) &= 1+\frac{1}{2}\int \pi (\bar\ell_w)^2 + \int\pi \sum_{k=3}^\infty \frac{1}{k!}(\bar\ell_w)^{k-2(1-\eps)}(\bar\ell_w)^{2(1-\eps)}\\
&\leq 1+\frac{1}{2}\int \pi (\bar\ell_w)^2 + \int\pi \sum_{k=3}^\infty \frac{1}{k!}(\eps|\ell'|+\delta)^{k-2(1-\eps)}|\bar\ell_w|^{2(1-\eps)}\\
&= 1+\frac{1}{2}\int \pi (\bar\ell_w)^2 + \int\pi \lt(\frac{e^{\eps|\ell'|+\delta}-1-(\eps|\ell'|+\delta) - \frac{1}{2}(\eps|\ell'|+\delta)^2}{(\eps|\ell'|+\delta)^{2(1-\eps)}}\rt)|\bar\ell_w|^{2(1-\eps)}\\
&\leq 1+\frac{1}{2}\int \pi (\bar\ell_w)^2 + \int\pi e^{\eps|\ell'|+\delta}|\bar\ell_w|^{2(1-\eps)}\\
&= 1+\frac{1}{2}\int \pi (\bar\ell_w)^2 + \frac{\int\pi_0 e^{(1-\eps)\ell'+\delta}|\bar\ell_w|^{2(1-\eps)}}{\int \pi_0e^{\ell'}}\\
&\leq 1+\frac{1}{2}\int \pi (\bar\ell_w)^2 + e^{\delta}\frac{\lt(\int\pi_0 e^{\ell'}|\bar\ell_w|^{2}\rt)^{1-\eps}}{\int \pi_0e^{\ell'}}\\
&= 1+\frac{1}{2}\int \pi (\bar\ell_w)^2 + e^{\delta}\lt(\int\pi_0 e^{\ell'}\rt)^{-\eps}\lt(\int \pi (\bar\ell_w)^2\rt)^{1-\eps}\\
&= 1+\frac{1}{2}\int \pi (\bar\ell_w)^2 + e^{\delta+c\eps}\lt(\int\pi_0 e^{\ell}\rt)^{-\eps}\lt(\int \pi (\bar\ell_w)^2\rt)^{1-\eps}\\
&\leq \exp\lt(h( w^T \Cov_\pi(\ell) w)\rt), 
\]
where $h(x) = \frac{1}{2}x + \frac{e^{\delta+c\eps}}{\int\pi_0 e^{\eps\ell}} x^{1-\eps}$.
\eprfof

\bprop\label{prop:upperrecovered2}
Suppose $\Theta = \reals^d$, $\bar\ell$ is $G$-strongly concave, and there exists $L<G$, $\alpha > 0$, and $\theta_0\in\Theta$ such that for 
all coreset weights $w$ satisfying $\alpha w^T\Cov_\pi((\ell_n)_{n=1}^N)w \leq 1$, $\bar\ell_w$ is $L$-Lipschitz smooth,
and both $\|\grad\ell_w(\theta_0)\|$ and $\bar\ell_w(\theta_0)$ are bounded.
Then for any $(L/G) < \eps < 1$, there exists $c\in\reals$, $\delta > 0$ such that the potentials $(\ell_n)_{n=1}^N$ 
are $(h, \alpha\Cov_\pi((\ell_n)_{n=1}^N))$-subexponential
with the same $h$ as in \cref{prop:upperrecovered}.
\eprop

\bprfof{\cref{prop:upperrecovered2}}
Since  $\bar\ell$ is $G$-strongly concave and $\bar\ell_w$ is $L$-Lipschitz smooth, we can write 
\[
\ell(\theta) &\leq \ell(\theta_0) + \grad\ell(\theta_0)^T(\theta-\theta_0)-\frac{G}{2}\lt\|\theta-\theta_0\rt\|^2\\
&= \ell(\theta_0) + \frac{G}{2}\lt\|G^{-1}\grad\ell(\theta_0)\rt\|^2 - \frac{G}{2}\lt\|\theta-\theta_0 - G^{-1}\grad\ell(\theta_0)\rt\|^2\\
|\bar\ell_w(\theta)| &\leq |\bar\ell_w(\theta_0) + \grad\ell_w(\theta_0)^T(\theta-\theta_0)| + \frac{L}{2}\|\theta-\theta_0\|^2.
\]
So setting $c = \ell(\theta_0) + \frac{G}{2}\lt\|G^{-1}\grad\ell(\theta_0)\rt\|^2$ implies $\ell-c$ is a nonpositive function as required.
Then
\[
|\bar\ell_w(\theta)| - \eps|\ell(\theta)-c| &\leq 
|\bar\ell_w(\theta_0)| + \frac{\eps}{2G}\|\grad\ell(\theta_0)\|^2+ \lt(\|\grad\ell_w(\theta_0)\| +\eps\|\grad\ell(\theta_0)\|\rt) \|\theta-\theta_0\|
+ \frac{L-\eps G}{2}\|\theta-\theta_0\|^2 .
\]
For $0 < a < G-L$, setting $\epsilon = \frac{L+a}{G}$ and then maximizing over $\|\theta-\theta_0\|$ yields
\[
|\bar\ell_w(\theta)| - \eps|\ell(\theta)-c| &\leq 
|\bar\ell_w(\theta_0)| + \frac{\eps}{2G}\|\grad\ell(\theta_0)\|^2+ \lt(\|\grad\ell_w(\theta_0)\| +\eps\|\grad\ell(\theta_0)\|\rt) \|\theta-\theta_0\|
- \frac{a}{2}\|\theta-\theta_0\|^2 \\
 &\leq 
|\bar\ell_w(\theta_0)| + \frac{\eps}{2G}\|\grad\ell(\theta_0)\|^2+ \frac{\lt(\|\grad\ell_w(\theta_0)\| +\eps\|\grad\ell(\theta_0)\|\rt)^2}{2a}.
\]
By the boundedness of $\bar\ell_w(\theta_0)$ and $\grad\ell_w(\theta_0)$, maximizing over $w$ yields a value of $\delta < \infty$.
\eprfof

\blem\label{lem:resamplingclt}
Let $X_1, X_2, \dots$ be \iid random variables in $\reals$ with $\E X_n = 0$,
and define the resampled sum
\[
S_{N} = \sum_{n=1}^N \frac{M_n}{Mp_n} X_n
\]
where $(M_1, \dots, M_N) \dist \Multi(M, (p_1,\dots, p_N))$,
with strictly positive resampling probabilities $p_1,\dots,p_N$ that may depend on $X_1, \dots, X_N$ and $N$.
If there exists a $\delta > 0$ such that as $N\to\infty$,
\[
\frac{1}{N}\sum_{n} \frac{|X_n|^{2+\delta}}{(Np_n)^{1+\delta}} &= O_p(1), \quad
\frac{1}{N}\sum_{n} \frac{X_n^2}{Np_n} = \Omega_p(1), \quad\text{and}\quad 
M\to \infty,
\]
then
\[
\sqrt{M}\frac{\frac{1}{N}S_N - \frac{1}{N}\sum_n X_n}{\sqrt{\frac{1}{N}\sum_n \frac{X^2_n}{Np_n}}} \convd \Norm(0,1).
\]
\elem
\bprf
We can rewrite 
\[
S_N = \frac{1}{M} \sum_{m=1}^{M} \frac{X_{I_m}}{p_{I_m}}
\]
where $I_m \distiid \Cat(p_1,\dots,p_N)$.
Consider $S_N + B_N$ where $B_N$ is independent of $S_N$,  $B_N = \pm 1$ with probability $\frac{1}{2(NM)^{1+\delta}}$, 
and $B_N = 0$ otherwise.
So if we set $\scA_N = \sigma(X_1,\dots,X_N)$,
\cite[Cor.~3]{Bulinski17} asserts that
\[
\frac{S_N+B_N - \E\left[S_N| \scA_N\right]}{\sqrt{(NM)^{-(1+\delta)}+\Var\left[S_N | \scA_N\right]}} \convd \Norm(0,1) \qquad N\to\infty.
\]
as long as for all $N$ large enough,
\[
\Var\left[\frac{1}{M}\frac{X_{I_m}}{p_{I_m}}| \scA_N\right] &< \infty \quad \text{\as},
\]
and as $N\to\infty$,
\[
\frac{
(NM)^{-(1+\delta)}
+\sum_{m=1}^{M}
\E\left[\left|\frac{1}{M}\frac{X_{I_m}}{p_{I_m}} - 
\E\left[\frac{1}{M}\frac{X_{I_m}}{p_{I_m}}| \scA_N\right]\right|^{2+\delta}
| \scA_N\right]
}{\left((NM)^{-(1+\delta)}+\Var\left[S_N | \scA_N\right]\right)^{(2+\delta)/2}} 
\convp 0.
\]
Note that the conditional mean and variance have the form
\[
\E\left[S_N | \scA_N\right] &= \E\left[\frac{X_{I_m}}{p_{I_m}} | \scA_N\right] = \sum_{n} X_n\\
\Var\left[S_N | \scA_N\right] &= \frac{1}{M} \Var\left[ \frac{X_{I_m}}{p_{I_m}}|\scA_N\right] 
=\frac{1}{M}\sum_{n} p_n\left(\frac{X_n}{p_n} - \sum_{n}X_n\right)^2,
\]
which implies that
$\Var\left[\frac{1}{M}\frac{X_{I_m}}{p_{I_m}}| \scA_N\right] < \infty$ \as,
since $p_1,\dots,p_N$ are strictly nonnegative and
$\E X_n = 0$ implies $X_n$ is finite almost surely.
Next, note that 
\[
&\frac{
(NM)^{-(1+\delta)}
+\sum_{m=1}^{M}
\E\left[\left|\frac{1}{M}\frac{X_{I_m}}{p_{I_m}} - 
\E\left[\frac{1}{M}\frac{X_{I_m}}{p_{I_m}}| \scA_N\right]\right|^{2+\delta}
| \scA_N\right]
}{\left((NM)^{-(1+\delta)}+\Var\left[S_N | \scA_N\right]\right)^{(2+\delta)/2}} \\
\leq&
\frac{
(NM)^{-(1+\delta)}
+2^{2+\delta}\sum_{m=1}^{M}
\left(\E\left[\left|\frac{1}{M}\frac{X_{I_m}}{p_{I_m}}\right|^{2+\delta}| \scA_N\right] + 
\left|\frac{1}{M}\sum_n X_n\right|^{2+\delta}\right)
}{\left((NM)^{-(1+\delta)}+\Var\left[S_N | \scA_N\right]\right)^{(2+\delta)/2}} \\
=&
M^{-\delta/2}
\frac{
N^{-(3+2\delta)}+2^{2+\delta}
\left(\frac{1}{N}\sum_n\frac{|X_n|^{2+\delta}}{(N p_n)^{1+\delta}} + 
\left|\frac{1}{N}\sum_n X_n\right|^{2+\delta}\right)
}{\left(M^{-\delta} N^{-(3+\delta)}+\frac{1}{N}\sum_{n} \frac{X_n^2}{ Np_n} - \left(\frac{1}{N}\sum_{n}X_n\right)^2\right)^{(2+\delta)/2}}.
\]
The above expression converges in probability to 0
by the technical assumptions in the statement of the result
as well as the fact that
$\frac{1}{N}\sum_{n} X_n \convas 0$ by the law of large numbers.
Once again by the technical assumptions,
$\Var\left[S_N | \scA_N\right] = \Omega_p(N^2/M)$,
so
\[
\frac{\Var\left[S_N| \scA_N\right]}{(NM)^{-(1+\delta)} + \Var\left[S_N| \scA_N\right]} &\convp 1\\
\frac{B_N}{(NM)^{-(1+\delta)} + \Var\left[S_N| \scA_N\right]} &\convp 0,
\]
and hence by Slutsky's theorem,
\[
\frac{S_N- \E\left[S_N| \scA_N\right]}{\sqrt{\Var\left[S_N | \scA_N\right]}} \convd \Norm(0,1) \qquad N\to\infty.
\]
Using Slutsky's theorem again with $\frac{1}{N}\sum_n X_n \convp 0$ and rearranging yields the final result.
\eprf

\blem\label{lem:Gconvp}
Suppose coreset weights are generated using the importance weighted construction in \cref{alg:subsampling}. 
Let $g = \grad\ell(\eta_0)$, $g_w = \grad\ell_w(\eta_0)$, and $H = -\E\lt[\grad^2\ell_n(\eta_0)\rt]$.
If conditions A(1-3) and (A6) in \cref{sec:applications_lower} hold, $M = o(N)$, and $M = \omega(1)$, then
\[
\lt\|\frac{g}{N}\rt\|_2 = \Theta_p\lt(N^{-1/2}\rt), \qquad \lt\|\frac{g_w}{\sbw}\rt\|_2 = \Theta_p\left(M^{-1/2}\right), \qquad \frac{\sbw}{N}\convp 1,
\]
and 
\[
\sup_{\|\eta-\eta_0\|_2\leq r} \left\|-\frac{1}{N}\grad^2\ell(\eta) - H\right\|_2 &\convp 0, &  \sup_{\|\eta-\eta_0\|_2\leq r} \left\|-\frac{1}{\sbw}\grad^2\ell_w(\eta) - H\right\|_2 &\convp 0.
\]
\elem
\bprf
First, since $\sbw = \sum_n \frac{M_n}{Mp_n}$, $\E\sbw = N$, and
\[
\E\lt[(\sbw - N)^2\rt] &= \frac{N^2}{M^2}\E\lt[\lt(\sum_n M_n\lt((Np_n)^{-1}-1\rt) \rt)^2 \rt]\\
&= \frac{N^2}{M^2}\lt( \sum_n ((Np_n)^{-1}-1)^2 \E M_n^2 + \sum_{n\neq n'} ((Np_n)^{-1}-1)((Np_{n'})^{-1}-1)\E[M_nM_{n'}]\rt)\\
&= \frac{N^2}{M}\lt( \sum_n ((Np_n)^{-1}-1)^2p_n - \lt(\sum_{n'} (1/N-p_n)\rt)^2\rt)\\
&= \frac{1}{M}\lt( \sum_n p_n(p_n^{-1}-N)^2\rt)\\
&\leq \frac{1}{M}\lt( \max_n (p_n^{-1}-N)^2\rt)\\
&\leq \frac{N^2}{M} O(1),
\]
where the last line follows by assumption A6. Therefore by Chebyshev's inequality and $M \to \infty$,
$\sbw / N \convp 1$.
Since the data are \iid, by conditions A1 and A2, the central limit theorem holds for the sum of $\grad\ell_n(\eta_0)$ 
such that $g/\sqrt{N}$ converges in distribution to a normal, 
and hence $\lt\|\frac{g}{N}\rt\| = \Theta_p(N^{-1/2})$.
By conditions A1, A2, and A6, \cref{lem:resamplingclt} holds such that for any $t\in\reals^d$,
\[
\sqrt{M} \frac{\frac{1}{N}t^Tg_w - \frac{1}{N}t^Tg}{\sqrt{\frac{1}{N}\sum_n \frac{(t^T\grad\ell_n(\eta_0))^2}{Np_n}}} = \Theta_p(1).
\]
Since condition A6 asserts that $C > Np_n \geq c > 0$, the law of large numbers, condition A1, and $M/N \to 0$ imply that
\[
\frac{\sqrt{M}}{N}t^Tg_w  = \Theta_p(1).
\]
Summing over a basis of vectors $t_1, \dots, t_d$ shows that
\[
\frac{\sqrt{M}}{N}\|g_w\|_2  = \Theta(1)\sqrt{M}\lt\|\frac{g_w}{\sbw}\rt\|_2 \Theta_p(1).
\]
This completes the first three results.
Next, by condition A3, for sufficiently large $N$ such that the neighbourhood contains the ball of radius $r$ around $\eta_0$,
\[
\sup_{\|\eta-\eta_0\|_2\leq r}\left\|\frac{1}{N}\grad^2\ell(\eta) - \frac{1}{N}\grad^2\ell(\eta_0)\right\|_2 &\leq r\frac{1}{N}\sum_n R(X_n)\\
\sup_{\|\eta-\eta_0\|_2\leq r}\left\|\frac{1}{N}\grad^2\ell_w(\eta) - \frac{1}{N}\grad^2\ell_w(\eta_0)\right\|_2 &\leq r\frac{1}{N}\sum_n w_n R(X_n),
\]
and
\[
\E\left[r\frac{1}{N}\sum_n R(X_n)\right] = \E\left[r\frac{1}{N}\sum_n w_n R(X_n)\right] = r\E\left[R(X)\right] \to 0,
\]
so that we have that both 
\[
\sup_{\|\eta-\eta_0\|_2\leq r}\left\|\frac{1}{N}\grad^2\ell(\eta) - \frac{1}{N}\grad^2\ell(\eta_0)\right\|_2 \convp 0
\quad\text{and}\quad
\sup_{\|\eta-\eta_0\|_2\leq r}\left\|\frac{1}{N}\grad^2\ell_w(\eta) - \frac{1}{N}\grad^2\ell_w(\eta_0)\right\|_2 \convp 0
\]
by Markov's inequality. Finally, by the bounded variance in A2, sampling probability bounds in A6, and $M\to\infty$, the variances of
$\frac{1}{N}\grad^2\ell_w(\eta_0)$ and $\frac{1}{N}\grad^2\ell(\eta_0)$ both converge to 0 as $N \to \infty$,
and since both of these quantities are unbiased estimates of $\E\lt[\grad^2\ell_n(\eta_0)\rt]$, Chebyshev's inequality
yields the desired convergence in probability.
\eprf

\blem\label{lem:vectorconc}
Suppose $(X_n)_{n=1}^N$ are  $N$ \iid random vectors in $\reals^d$.
Fix $M \in \nats$, $M< d$ and define $X = \bbmat X_1 & X_2 & \dots & X_M\ebmat \in \reals^{d\times M}$.
If there exists $\delta > 0$ such that
\[
\E\lt[ (1^T(X^TX)^{-1}1)^{M+\delta} \rt] < \infty,
\]
where $1$ denotes a vector of all 1 entries,
then as $N\to\infty$,
\[
\lt(\begin{aligned}
\min_{w \in \reals^N_+} &\lt\|\frac{\sum_{n=1}^N w_n X_n}{\sum_{n=1}^N w_n}\rt\|^2\\
\st & \sum_n \1[w_n > 0] < M.
\end{aligned}\rt) = \omega_p\lt(N^{-\frac{M+\delta/2}{M+\delta}}\rt).
\]
\elem
\bprf
For any $\epsilon > 0$, by the union bound over subsets of $[N]$ of size $M$,
\[
\P\lt(  \min_{w\in\reals^N_+} \dots \leq \eps\rt) &\leq {N \choose M} \P\lt( \min_{w \in \reals^M} \frac{w^TX^TXw}{w^T11^Tw} \leq \eps\rt)\\
&\leq {N \choose M} \P\lt( \max_{\lambda} \min_{w \in \reals^M} w^TX^TXw -\lambda(1^Tw - 1) \leq \eps\rt)\\
&= {N \choose M} \P\lt( \max_{\lambda} \lambda  - \frac{\lambda^2}{4} 1^T(X^TX)^{-1}1  \leq \eps\rt)\\
&= {N \choose M} \P\lt(  1^T(X^TX)^{-1}1 \geq \eps^{-1}\rt).
\]
By Markov's inequality and ${N\choose M} \leq (eN/M)^M$,
\[
\P\lt(  \min_{w\in\reals^N_+} \dots \leq \eps\rt)
&\leq \lt(\frac{eN}{M}\rt)^M \eps^{M+\delta}\E\lt[(1^T(X^TX)^{-1}1)^{M+\delta}\rt]\\
&= \lt(\frac{eN\eps^{\frac{M+\delta}{M}}}{M}\rt)^M \E\lt[(1^T(X^TX)^{-1}1)^{M+\delta}\rt].
\]
Setting $\epsilon = N^{-\frac{M+\delta/2}{M+\delta}}$ yields
\[
\P\lt(  \min_{w\in\reals^N_+} \dots \leq N^{-\frac{M+\delta/2}{M+\delta}}\rt)
&\leq \lt(\frac{eN^{- \frac{\delta}{2M}}}{M}\rt)^M \E\lt[(1^T(X^TX)^{-1}1)^{M+\delta}\rt].
\]
The right-hand side converges to 0 as $N\to\infty$, yielding the stated result.
\eprf

\bprfof{\cref{cor:importanceweighted} and \cref{cor:scaledimportanceweighted}}
Set $r = (\log M)^{-1/2}$. 
Then since $M=o(N)$, $M=\omega(1)$, and assumptions (A1-3) and (A6) hold, 
\cref{lem:Gconvp} holds. Note that $\|g_w/\sbw\| = \Theta_p(M^{-1/2}) = o_p(r)$,
$\eta\pi_0$ is positive at $\eta_0$ and twice differentiable by (A4),
and $Nr^2 = N/\log M = \omega(1)$ since $M=o(N)$.
Thus the conditions of \cref{thm:diffable_lower} are verified.
Substitution into the right term in the minimum of \cref{thm:diffable_lower} yields the stated lower bound of $\Omega_p(N/M)$.
For the left term in \cref{thm:diffable_lower}, define $B = \{(\eta-\eta_0)^TH(\eta-\eta_0) \leq r^2\}$.
Then since $H\succ 0$, $r\to 0$, and $r^2 = 1/\log M = \omega(\log N /N)$,
(A5) guarantees that $-\log (\eta\pi)(B^c) = \Omega_p(Nr^2) = \Omega_p(N/\log M)$.
Therefore the minimum is $\Omega_p(N/M)$, and
we complete the proof by transferring from $\KLmin(w)$ on the $\eta$-pushforward model to
$\KLmin(w)$ on the original model  using \cref{cor:lower_other}.
\eprfof

\bprfof{\cref{cor:highdimlowerbd}}
Fix the $\delta > 0$ guaranteed by (A8), and set $r = N^{-\frac{M+3\delta/4}{2(M+\delta)}}$.
Note that $Nr^2 = N^{\frac{\delta/4}{M+\delta}} = \omega(1)$,
$\eta\pi_0$ is positive at $\eta_0$ and twice differentiable by (A4),
and by (A1-3) the results pertaining to $\lt\|\frac{g}{N}\rt\|_2$ and $\sup_{\|\eta-\eta_0\|_2\leq r}\lt\|-\frac{1}{N}\grad^2\ell(\eta)-H\rt\|_2$
in \cref{lem:Gconvp} hold; thus \cref{assum:lower} holds.
By (A7), \cref{assum:lower2} holds as well as the conditions on $\frac{1}{\sbw}\grad^2\ell_w(\theta)$ 
and $\frac{1}{\sbw}\sum_n w_nL_n^2$ in \cref{thm:diffable_lower_min}. 
Finally by (A8), \cref{lem:vectorconc} holds such that
\[
\lt\|\frac{g_w}{\sbw}\rt\|^2 = \omega_p\lt(N^{-\frac{M+\delta/2}{M+\delta}}\rt),
\]
and hence $\lt\|\frac{g_w}{\sbw}\rt\| = \omega_p(r)$.
Therefore all conditions of \cref{thm:diffable_lower_min} hold.
For the left term in the minimum in \cref{thm:diffable_lower_min}, 
define $B = \{(\eta-\eta_0)^TH(\eta-\eta_0) \leq r^2\}$.
Then since $H\succ 0$, $r\to 0$, and $r^2 = N^{-\frac{M+3\delta/4}{M+\delta}} = \omega(\log N /N)$,
(A5) guarantees that $-\log (\eta\pi)(B^c) = \Omega_p(Nr^2) = \Omega_p\lt(N^{\frac{\delta/4}{M+\delta}}\rt)$.
For the right term,
\[
\log\lt(N\lt\|\frac{g_w}{\sbw}\rt\|^2\rt) = \Omega_p\lt(\log N^{1-\frac{M+\delta/2}{M+\delta}}\rt) = \Omega_p\lt(\log N\rt).
\]
The minimum of these two is from the right term, so
\[
\KLmin(w) = \Omega_p\lt(\log N\rt).
\]
We complete the proof by transferring from $\KLmin(w)$ on the $\eta$-pushforward model to
$\KLmin(w)$ on the original model  using \cref{cor:lower_other}.
\eprfof

\bprop\label{prop:cauchylogrega15}
The models specified in \cref{eq:cauchymodel,eq:logregmodel} satisfy assumptions (A1-5).
\eprop
\bprf
The exact same technique applies to both models, so here we will just demonstrate it for the Cauchy
model. In the Cauchy model, $\theta\in\reals$, $\eta : \reals\to\reals_+$, $\eta(\theta) = \theta^2$, and
\[
\ell_n(\eta) &= -\log \pi - \log ((Z_n - \eta)^2+1) &
\grad\ell_n(\eta) &= \frac{2(Z_n-\eta)}{(Z_n-\eta)^2+1} \\
\grad^2\ell_n(\eta) &= \frac{2(Z_n-\eta)^2-2}{((Z_n-\eta)^2+1)^2} &
\grad^3\ell_n(\eta) &= \frac{4((Z_n-\eta)^2-3)(\eta-Z_n)}{((\eta-Z_n)^2+1)^3},
\]
where $Z_n = X_n$. Property (A1) holds by routine interchange of differentiation and integration.
Property (A2) holds (for any $\delta > 0$) because  $\grad\ell_n(\eta)$ and $\grad^2\ell_n(\eta)$ are 
bounded functions of $\eta$ and $X_n$ jointly.
Property (A3) holds (for any neighbourhood of $\eta_0$)
because $\grad^3\ell_n(\eta)$ is a bounded function of $\eta$ and $X_n$ jointly.
Property (A4) holds because the pushforward of $\Cauchy(0, 1)$ through $\eta(\theta) = \theta^2$
has full support on $\reals_+$.
In order to verify assumption (A5),
suppose there 
exists a sequence of bounded measurable functions $\phi_r(Z_1, \dots, Z_N) \in [0,1]$ of the data
and constants $c,c' > 0$ such that for all $r\to 0$, $r^2 = \omega(\log N / N)$,
\[
\E_{\eta_0}\phi_r = O\lt(e^{-cNr^2}\rt) \quad\text{and}\quad \sup_{\|\eta-\eta_0\| > r}\E_{\eta}(1-\phi_r) = O\lt(e^{-c'Nr^2}\rt).
\]
The functions $\phi_r$ are similar to the test functions of Schwartz \cite{Schwartz65}.
Then defining $\mu = \eta\pi$ and $\mu_0 = \eta\pi_0$,
\[
\mu(\|\eta-\eta_0\| > r) &= \phi_r \mu(\|\eta-\eta_0\| > r) + (1-\phi_r)\mu(\|\eta-\eta_0\| > r) \\
&\leq \phi_r + (1-\phi_r)\mu(\|\eta-\eta_0\|>r)\\
&= \phi_r + \frac{\int_{\|\eta-\eta_0\|>r} (1-\phi_r)e^{\ell(\eta)-\ell(\eta_0)} \mu_0 }{\int e^{\ell(\eta)-\ell(\eta_0)} \mu_0}.
\]
Using the same proof technique as in \cref{thm:diffable_lower}, the denominator satisfies
\[
\log\int e^{\ell(\eta)-\ell(\eta_0)} \mu_0(\d\eta) \geq -\frac{d}{2}\log N + O_p(1).
\]
By assumption, there exists $c>0$ such that
\[
\E_{\eta_0}[\phi_r] = O\lt(e^{-c Nr^2}\rt) \implies \phi_r = O_p(e^{-cNr^2}),
\]
and a $c'>0$ such that
\[
\E_{\eta_0}\lt[\int_{\|\eta-\eta_0\|>r} (1-\phi_r)e^{\ell(\eta)-\ell(\eta_0)} \mu_0\rt]
&= \int_{\|\eta-\eta_0\|>r} \E_{\eta}(1-\phi_r)\mu_0\\
&\leq \sup_{\|\eta-\eta_0\|>r} \E_{\eta}(1-\phi_r)  \\
&= O(e^{-cNr^2}) \implies \int_{\|\eta-\eta_0\|>r}(\dots) = O_p\lt(e^{-cNr^2}\rt).
\]
Therefore $\mu(\|\eta-\eta_0\|\geq r) = O_p\lt( e^{-cN r^2} + N^{d/2} e^{-c'Nr^2}\rt) = O_p\lt(e^{(d/2)\log N - c''Nr^2}\rt)$; and 
since $r^2 = \omega(\log N/N)$, $-\log \mu(\|\eta-\eta_0\|\geq r) = \Omega_p( Nr^2)$ as required by (A5).
So to complete the proof of (A5) we need to find a suitable $\phi_r$. 
Fix $\eps > 0$, and set
\[
\phi_r(Z_1, \dots, Z_N) = \1\lt[P_{\eta_0}(|Z-\eta_0|\leq 1) - \frac{1}{N}\sum_{n=1}^N \1[|Z_n-\eta_0| \leq 1] > \eps r\rt].
\]
Under $p_{\eta_0}$, Hoeffding's inequality yields
\[
\E_{\eta_0}\phi_r \leq e^{-2N\eps^2 r^2}.
\]
And under $p_\eta$ for $\|\eta-\eta_0\| > r$, for small enough $\eps > 0$, $P_{\eta_0}(|Z-\eta_0|\leq 1) - P_{\eta}(|Z-\eta_0|\leq 1) \geq 2\eps r$.
Therefore
\[
\E_{\eta}\lt[1-\phi_r(Z_1, \dots, Z_N)\rt] &= \Pr\nolimits_\eta\lt(P_{\eta_0}(|Z-\eta_0|\leq 1) -\frac{1}{N}\sum_{n=1}^N \1[|Z_n-\eta_0| \leq 1] \leq \eps r\rt)\\
&\leq \Pr\nolimits_\eta\lt(P_\eta(|Z-\eta_0|\leq 1)  - \frac{1}{N}\sum_{n=1}^N \1[|Z_n-\eta_0| \leq 1] \leq -\eps r\rt)\\
&= \Pr\nolimits_\eta\lt(\frac{1}{N}\sum_{n=1}^N \1[|Z_n-\eta_0| \leq 1] - P_\eta(|Z-\eta_0|\leq 1) \geq \eps r\rt),
\]
at which point we can again apply Hoeffding's inequality, completing the result.
\eprf

\blem\label{lem:iterativeconebound}
Fix vectors $u, u_1, \dots, u_N$ in a separable Hilbert space with inner product denoted $a\cdot b$ and norm denoted $\|\,\|$.
Let $v_1,\dots,v_M$ be drawn from $\{u_1,\dots,u_N\}$ with probabilities $p_1,\dots,p_N$ either with or without replacement
(if without replacement, the probabilities are renormalized after every draw). Then
for all $\epsilon \geq 0$,
\[
\P\lt(\min_{w\geq 0}\lt\|\sum_{m=1}^M w_m v_m - u \rt\|^2 > \epsilon^{M\lt(\frac{q(M,\eps)}{2}\rt)+1}\|u\|^2\rt)
&\leq e^{-\lt(\frac{1-\log(2)}{2}\rt)M},
\]
where
\[
q(M,\eps) &= \P\lt( 1-\max\lt\{0, \frac{v_{M}}{\|v_M\|}\cdot\frac{(u-x_{M-1})}{\|u-x_{M-1}\|}\rt\}^2 \leq \eps\rt) & x_{M-1} &= \argmin_{x\in\cone\{v_1, \dots, v_{M-1}\}}\lt\|x - u\rt\|^2.
\]
\elem
\bprf
First note that it suffices to analyze the case with replacement, since this case provides an upper bound on the case without replacement. 
To demonstrate this,
we couple two probability spaces---one that draws $v_1,\dots,v_M$ with replacement, and one without replacement. First, 
draw an identical vector $v_1$ for both copies. On each subsequent iteration $m > 1$, the ``with replacement'' copy
first draws whether or not it selects a vector that was previously selected by the ``without replacement'' copy.
If it does, it draws that vector independently; if it does not, it selects the same vector as the ``without replacement'' copy.
In any case, at each iteration $m$, the vectors drawn by the ``with replacement'' copy are always a subset of the vectors drawn by the ``without replacement''
copy, and hence the minimum over $w\geq 0$ is greater for that copy. It therefore suffices to analyze the case with replacement.

To obtain an upper bound on the probability when sampling with replacement, instead of minimizing over all $w\geq 0$ jointly, suppose we use the following iterative algorithm.
Set $x_0 = 0$. At the first iteration, we draw $v_1$ and set the weight $w_1$ by optimizing over $w_1 \geq 0$:
\[
\min_{w_1 > 0} \lt\| w_1 v_1 - u\rt\|^2 = \|u\|^2\lt(1 - \max\lt\{0, \frac{v_1 \cdot u}{\|v_1\|\|u\|}\rt\}^2\rt).
\]
Set $x_1 = w_1v_1$, and note that $(u-x_1)\cdot x_1 = 0$. Then 
at each subsequent iteration $k$, assume the previous iterate is optimized over all nonnegative weights, 
and hence satisfies $(u-x_{k-1})\cdot x_{k-1} = 0$.
We draw another vector $v_k$, and bound the erorr of the next iterate $x_k$ by optimizing 
over only the weight $w_k$ for the new vector $v_k$.
Then 
\[
\|u-x_k\|^2 = \min_{w_1,\dots,w_k \geq 0} \lt\|\sum_{m=1}^k w_mv_m - u\rt\|^2 &\leq  \min_{w_k > 0} \lt\|w_k v_k + x_{k-1} - u\rt\|^2\\
&= \|u-x_{k-1}\|^2 \lt(1 - \max\lt\{0, \frac{v_k\cdot (u-x_{k-1})}{\|v_k\|\|u-x_{k-1}\|}\rt\}^2\rt).
\]
Therefore,
\[
&\hspace{0cm}\P\lt(\min_{w\geq 0}\lt\|\sum_{m=1}^M w_m v_m - u \rt\|^2 \leq \epsilon^K\|u\|^2\rt)\\
&\geq 
\P\lt(\text{in at least $K$ iterations, }\|x_k - u\|^2 \leq \eps \|x_{k-1}-u\|^2\rt)\\
&\geq 
\P\lt(\text{in at least $K$ iterations, } 1-\max\lt\{0, \frac{v_k\cdot(u-x_{k-1})}{\|v_k\|\|u-x_{k-1}\|}\rt\}^2 \leq \eps\rt)\\
&= \sum_{\scK\subseteq [M], |\scK| \geq K} \P\lt(k\in\scK \iff 1-\max\lt\{0, \frac{v_k\cdot(u-x_{k-1})}{\|v_k\|\|u-x_{k-1}\|}\rt\}^2 \leq \eps \rt)\\
&\geq \sum_{\scK\subseteq [M], |\scK| \geq K} q^k(1-q)^{M-k}\\
&= \sum_{k=K}^M {M\choose k} q^k(1-q)^{M-k},
\]
where
\[
q &= \P\lt( 1-\max\lt\{0, \frac{v_{M}\cdot(u-x_{M-1})}{\|v_{M}\|\|u-x_{M-1}\|}\rt\}^2 \leq \eps\rt)\\
x_{M-1} &= \argmin_{x\in\cone\{v_1, \dots, v_{M-1}\}}\lt\|x - u\rt\|^2
\]

So for all $0 \leq K \leq M$,
\[
\P\lt(\min_{w\geq 0}\lt\|\sum_{m=1}^M w_m v_m - u \rt\|^2 > \epsilon^K\|u\|^2\rt)
 &\leq \Binom(M, K-1, q).
 \]
Using the Chernoff bound on the binomial CDF, for all $K-1 \leq Mq$,
\[
\P\lt(\min_{w\geq 0}\lt\|\sum_{m=1}^M w_m v_m - u \rt\|^2 > \epsilon^K\|u\|^2\rt)
&\leq e^{-M\lt( \frac{K-1}{M}\log\frac{K-1}{Mq} + (1-\frac{K-1}{M})\log\frac{1-\frac{K-1}{M}}{1-q}\rt)}\\
&= e^{-(K-1)\log\frac{K-1}{Mq} - (M-(K-1))\log\frac{M-(K-1)}{M(1-q)}}.
\]
Substituting $K-1 = Mq/2$ yields
\[
= e^{M((q/2)\log 2 - (1-q/2)\log\frac{1-q/2}{(1-q)})} \leq e^{-\lt(\frac{1-\log(2)}{2}\rt)M}.
\]

\eprf

\bprfof{\cref{cor:subopt}}
Since the potentials are $\beta\Cov_\pi((\ell_n)_{n=1}^N)$ subexponential,
\cref{thm:subexpupper} guarantees that
\[
\forall w\in\reals_+^N : \, 4\beta (w-1)^T\Cov_\pi((\ell_n)_{n=1}^N)(w-1) \leq 1, \qquad \KLmax(w) \leq 4\beta (w-1)^T\Cov_\pi((\ell_n)_{n=1}^N)(w-1).
\]
We apply \cref{lem:iterativeconebound} with vectors $\ell_1,\dots,\ell_N$ (in equivalence classes specified up to a additive constant)
and inner product between $\ell_i, \ell_j$ defined by $\Cov_{\pi}(\ell_i, \ell_j)$.
In the notation of \cref{lem:iterativeconebound}, by assumption, $\|u\|^2 = O_p(N^\alpha)$ and $q(M,\eps) = \omega_p(M^{-\rho})$.
Substituting $M = (\log N)^{1/(1-\rho)}$, we find that 
\[
\P\lt( 4\beta (w-1)^T\Cov_\pi((\ell_n)_{n=1}^N)(w-1) \geq \eps^{-\omega_p(\log N) + \alpha \log N}\rt) \to 0.
\]
Combining this result with the KL bound above yields the final result.
\eprfof

\newpage
\section*{NeurIPS Paper Checklist}
\begin{enumerate}

\item {\bf Claims}
    \item[] Question: Do the main claims made in the abstract and introduction accurately reflect the paper's contributions and scope?
    \item[] Answer: \answerYes{} 
    \item[] Justification: The abstract and introduction state that the paper produces new general upper and lower bounds for Bayesian coreset approximations,
    that these bounds are applied to particular cases of interest, and that empirical results align with the theory. The paper does indeed contain these results, with proofs in the supplemental material,
    and empirical results in the figures do match the theory well.

\item {\bf Limitations}
    \item[] Question: Does the paper discuss the limitations of the work performed by the authors?
    \item[] Answer: \answerYes{} 
	\item[] Justification: In the conclusions section of the work, two main
	limitations of the theory are mentioned: the need to perform case-by-case
	analysis of subexponentiality constants and alignment probabilities in
	\cref{def:subexp} and \cref{cor:subopt}.  

\item {\bf Theory Assumptions and Proofs}
    \item[] Question: For each theoretical result, does the paper provide the full set of assumptions and a complete (and correct) proof?
    \item[] Answer: \answerYes{} 
    \item[] Justification: Theoretical results are numbered, and proofs of all results are included in the appendix.

    \item {\bf Experimental Result Reproducibility}
    \item[] Question: Does the paper fully disclose all the information needed to reproduce the main experimental results of the paper to the extent that it affects the main claims and/or conclusions of the paper (regardless of whether the code and data are provided or not)?
    \item[] Answer: \answerYes{} 
    \item[] Justification: All details needed to reproduce the experimental results in Figure 2 and 3 are included in the text. 
    Algorithms used in the experiments exist in the cited literature.

\item {\bf Open access to data and code}
    \item[] Question: Does the paper provide open access to the data and code, with sufficient instructions to faithfully reproduce the main experimental results, as described in supplemental material?
    \item[] Answer: \answerNo{} 
    \item[] Justification: There are no new algorithms presented in this work; the experiments involve only existing methods for which public code is available. The code is not 
    central to the contributions of the paper.

\item {\bf Experimental Setting/Details}
    \item[] Question: Does the paper specify all the training and test details (e.g., data splits, hyperparameters, how they were chosen, type of optimizer, etc.) necessary to understand the results?
    \item[] Answer: \answerYes{} 
    \item[] Justification: All details needed to reproduce the experimental results in Figure 2 and 3 are included in the text.
    Algorithms used in the experiments exist in the cited literature.

\item {\bf Experiment Statistical Significance}
    \item[] Question: Does the paper report error bars suitably and correctly defined or other appropriate information about the statistical significance of the experiments?
    \item[] Answer: \answerYes{}
    \item[] Justification: All empirical results show the mean over a number of trials, as well as error bars indicating standard error.

\item {\bf Experiments Compute Resources}
    \item[] Question: For each experiment, does the paper provide sufficient information on the computer resources (type of compute workers, memory, time of execution) needed to reproduce the experiments?
    \item[] Answer: \answerYes{}
    \item[] Justification: These details are not important for this paper, as there are no new methods or algorithms presented or claims related to computational performance. 
    However, the introduction does list that simulations were performed on a desktop computer with a Core i7 processor and 32GB RAM.

\item {\bf Code Of Ethics}
    \item[] Question: Does the research conducted in the paper conform, in every respect, with the NeurIPS Code of Ethics \url{https://neurips.cc/public/EthicsGuidelines}?
    \item[] Answer: \answerYes{} 
    \item[] Justification: This paper presents a new theoretical analysis of error bounds for Bayesian coresets methods. It does not present any new methodology or data with potential harmful consequences.

\item {\bf Broader Impacts}
    \item[] Question: Does the paper discuss both potential positive societal impacts and negative societal impacts of the work performed?
    \item[] Answer: \answerNA{}
    \item[] Justification: There is no potential negative societal impact of this work. The paper provides new theory regarding existing methodology.

\item {\bf Safeguards}
    \item[] Question: Does the paper describe safeguards that have been put in place for responsible release of data or models that have a high risk for misuse (e.g., pretrained language models, image generators, or scraped datasets)?
    \item[] Answer: \answerNA{}
    \item[] Justification: This does not apply.

\item {\bf Licenses for existing assets}
    \item[] Question: Are the creators or original owners of assets (e.g., code, data, models), used in the paper, properly credited and are the license and terms of use explicitly mentioned and properly respected?
    \item[] Answer: \answerNA{}
    \item[] Justification: This does not apply.

\item {\bf New Assets}
    \item[] Question: Are new assets introduced in the paper well documented and is the documentation provided alongside the assets?
    \item[] Answer: \answerNA{} 
    \item[] Justification: This does not apply.

\item {\bf Crowdsourcing and Research with Human Subjects}
    \item[] Question: For crowdsourcing experiments and research with human subjects, does the paper include the full text of instructions given to participants and screenshots, if applicable, as well as details about compensation (if any)?
    \item[] Answer: \answerNA{} 
    \item[] Justification: This does not apply.

\item {\bf Institutional Review Board (IRB) Approvals or Equivalent for Research with Human Subjects}
    \item[] Question: Does the paper describe potential risks incurred by study participants, whether such risks were disclosed to the subjects, and whether Institutional Review Board (IRB) approvals (or an equivalent approval/review based on the requirements of your country or institution) were obtained?
    \item[] Answer: \answerNA{} 
    \item[] Justification: This does not apply.

\end{enumerate}

\end{document}